\documentclass[journal]{IEEEtran}
\usepackage{amsmath}
\usepackage{makecell}
\usepackage{graphicx}
\usepackage{tabularx}
\usepackage{amssymb}
\usepackage{subcaption}
\usepackage{threeparttable}
\usepackage{amsmath,lipsum}
\usepackage{cuted}
\usepackage{cite}
\usepackage[square, comma, sort&compress, numbers]{natbib}
\usepackage{multirow}
\usepackage{color}
\usepackage{float}
\usepackage{stfloats}
\usepackage{booktabs}
\newcommand*{\red}{\textcolor{red}}
\usepackage{hyperref}
\newcommand*{\green}{\textcolor{green}}

\begin{document}
\title{ArrivalNet: Predicting City-wide Bus/Tram Arrival Time with Two-dimensional Temporal Variation Modeling}

\author{Zirui Li,~\IEEEmembership{Graduate Student Member,~IEEE,} Patrick Wolf, 
Meng Wang,~\IEEEmembership{Senior Member,~IEEE}        
\thanks{Zirui Li and Meng Wang are with the Chair of Traffic Process Automation, "Friedrich List" Faculty of Transport and Traffic Sciences, TU Dresden, 01069, Dresden, Germany. (Email: zirui.li@tu-dresden.de; meng.wang@tu-dresden.de)}
\thanks{Patrick Wolf is with the Traffic Management Department of the Dresden Transport Company, Dresdner Verkehrsbetriebe AG, 01129, Dresden, Germany. (Email: patrick.wolf@dvbag.de)}
\thanks{Corresponding author: Meng Wang.}
}


\markboth{}%
{Li \MakeLowercase{\textit{et al.}}: ArrivalNet: Predicting City-wide Bus/Tram Arrival Time with Two-dimensional Temporal Variation Modeling}

\maketitle

\begin{abstract}
Accurate arrival time prediction (ATP) of buses and trams plays a crucial role in public transport operations. Current methods focused on modeling one-dimensional temporal information but overlooked the latent periodic information within time series. Moreover, most studies developed algorithms for ATP based on a single or a few routes of public transport, which reduces the transferability of the prediction models and their applicability in public transport management systems. To this end, this paper proposes \textit{ArrivalNet}, a two-dimensional temporal variation-based multi-step ATP for buses and trams. It decomposes the one-dimensional temporal sequence into intra-periodic and inter-periodic variations, which can be recast into two-dimensional tensors (2D blocks). Each row of a tensor contains the time points within a period, and each column involves the time points at the same intra-periodic index across various periods. The transformed 2D blocks in different frequencies have an image-like feature representation that enables effective learning with computer vision backbones (e.g., convolutional neural network). Drawing on the concept of residual neural network, the 2D block module is designed as a basic module for flexible aggregation. Meanwhile, contextual factors like workdays, peak hours, and intersections, are also utilized in the augmented feature representation to improve the performance of prediction. 125 days of public transport ta from Dresden were collected for model training and validation. Experimental results show that the root mean square error, mean absolute error, and mean absolute percentage error of the proposed predictor decrease by at least 6.1\%, 14.7\%, and 34.2\% compared with state-of-the-art baseline methods. 
\end{abstract}


\section{Introduction}
\begin{figure*}[t]
  \centering
  \includegraphics[width=1\textwidth]{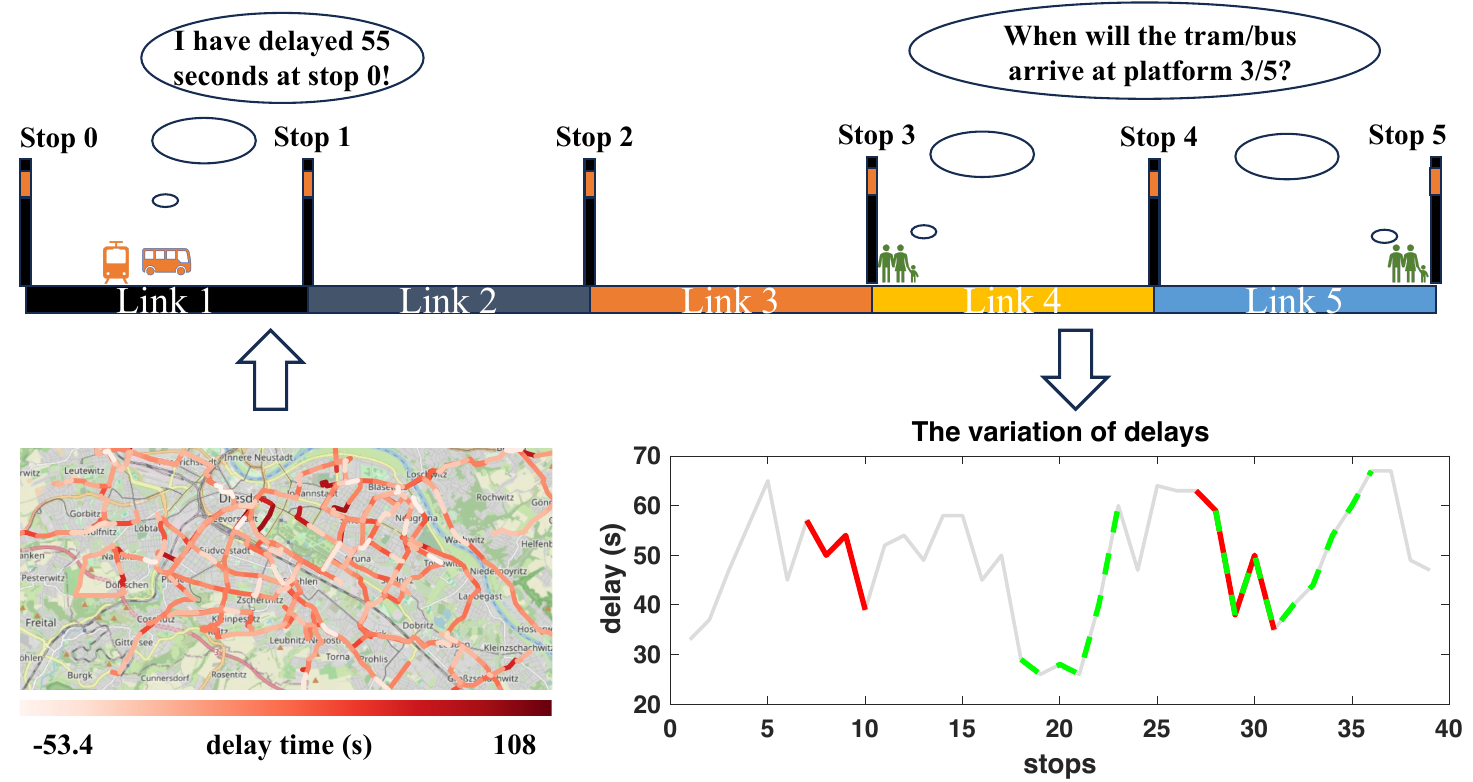}
  \caption{The problem illustration of multi-step bus/tram arrival time prediction. \textbf{Bottom left}: The city-wide public transport link delay distribution of tram in Dresden, Germany. (A link is the route segment between adjacent stops.) \textbf{Top}: When the bus/tram is on Link 1 (between stop 0 and stop 1) and has delayed 55 seconds, the potential passengers in the future stops are concerned on \textit{when the bus/tram will arrive at stop 3/5}?  \textbf{Bottom right}: The analysis from the two-dimensional perspective. In a sequence of tram delay information with 39 stops, the temporal pattern of \red{falling} and \green{fluctuation} are labelled in \red{red} and \green{green}, respectively.}
  \label{fig_wait}
\end{figure*}

As urban populations swell and individual car ownership rises, traffic congestion, energy consumption, and air pollution pose increasing challenges to urban transport systems. Public transport is one of the promising options to address the issues and achieve sustainable transport~\cite{hensher2007sustainable,buehler2011making}. However, existing public transport (PT) systems (buses, trams, subways, etc.) often suffer from issues of low reliability and prolonged delays in arrival times~\cite{soza2019underlying,amberg2019robust}.~\cite{lam2001value} indicated that users were significantly concerned with the accuracy of travel and arrival time predictions (ATP), which greatly influences their travel choices and experiences. Fig.~\ref{fig_wait} illustrates the multi-step bus/tram ATP. In this scenario, when the bus/tram is on Link 1 (between stop 0 and stop 1) and has been delayed 55 seconds, the passengers in the future stops would like to know \textbf{when will the bus/tram arrive at stops 3-5?} 

Accurate ATP not only facilitates the rescheduling and dispatching of public transport for operators but also helps passengers make informed decisions regarding their traveling plans and mode choices\cite{cheng2024analytical,zhou2022meso}. It needs to take into account of many influencing factors: the traffic situation (e.g., traffic volume and speed), platform facility design, passenger demand, the propagation of delay, traffic signals, etc~\cite{Ratneswaran,ma2019bus,kim2024computational,guarda2024estimating}. All elements above lead to the complexity and difficulty in the accurate ATP. To cope the issues, numerous solutions have been proposed to predict the arrival time~\cite{buchel2020review,singh2022review}, which can be divided into three categories: state estimation-based and statistical learning-based and deep learning-based methods. In the sequel, we survey the main methods in each category to identify the knowledge gaps.


State estimation-based methods treat the public transport operation process as a dynamic system and the kinematic state of public transport vehicles as the system state to be estimated. To this end, popular state estimation approaches of Kalman filter (KF) and Bayesian networks from the systems engineering domain can be applied for ATP~\cite{welch1995introduction,heckerman2008tutorial,AcharBus2020}. KF is an efficient state estimation method, as it can update the linear system state when new observations become available continuously.~\cite{cathey2003prescription} proposed a KF-based model for vehicle arrival/departure prediction using real-time and historical data. It makes optimal estimation of the location and speed of the vehicle based on the streaming data from the automatic vehicle location (AVL) system.~\cite{AcharBus2020} formulated a spatial KF to detect the unknown order of spatial dependence, and then learn its linear, non-stationary spatial correlations for this detected order. However, the performance is limited by the linear formulation of the state-space model. \cite{isukapati2020hierarchical} proposed a hierarchical Bayesian framework for bus dwell time prediction with minimal historical data. It made predictions using a small set of continually updated model parameter distributions, which was inherently adaptive to the time-varying duration of dwell time. The Bayesian architecture also provided the confidence in dwell time estimation to support the decision-making of scheduling under uncertainty. In~\cite{chen2023probabilistic}, a Bayesian Gaussian mixture model (GMM) was developed to generate probabilistic forecasting of bus travel time. It characterizes the strong dependencies between adjacent buses (e.g., correlated speed and smooth variation of headway). In the inference stage, an efficient Markov chain Monte Carlo (MCMC) algorithm was applied for probabilistic prediction. Furthermore, a conditional forecasting model for bus travel time and passenger occupancy was proposed with a similar strategy~\cite{chen2024conditional}. The advantage of state estimation-based methods is their ability to explicitly model the relationships among historical observations, noise, and uncertainties. Additionally, these methods stand out for their efficiency, requiring only a minimal dataset to operate effectively. However, the relationship between some factors and ATP is nonlinear (e.g., weather conditions), making it daunting to formulate these implicit and intertwined relationships explicitly. Meanwhile, for long-term predictions, the influence between nonadjacent samples and the propagation of delays in time series becomes the bottleneck of the state estimation-based methods~\cite{Ratneswaran}.

Statistical learning-based approaches for public transport ATP aim at capturing complex public transport modalities and nonlinear relationships inherent in the collected data from the perspective of statistics~\cite{ma2019bus,Ratneswaran}.~\cite{yang2016bus} integrated support vector machine (SVM) and genetic algorithm to search the best parameters in predicting the bus arrival time in various traffic conditions.~\cite{liu2012bus} proposed a \textit{k}-nearest neighbor (\textit{k}-NN)-based framework for bus arrival time estimation, which outperformed simple artificial neural networks (ANNs). 
In~\cite{yu2011bus}, ANN, KF, \textit{k}-NN, and linear regression (LR) were adopted for the bus ATP at the same bus stop but with different routes using real-world data. Compared with the bus information of the same route, multi-source from several routes can provide more benefit in the timeliness and reliability of the information. Meanwhile, the performance of several prediction methods are assessed and a valuable insight for algorithm selection was provided, which guided the further development of travel time predictors. In~\cite{ma2019bus}, various statistic machine learning algorithms were compared and evaluated. It demonstrated that real-time traffic information from taxis could improve the performance of ATP based on bus GPS data under both normal and abnormal traffic conditions.~\cite{kumar2018hybrid} integrated KF and \textit{k}-NN algorithm for bus travel time estimation. It applied \textit{k}-NN algorithm to capture the related input features and then combined exponential
smoothing technique with a recursive estimation scheme based on KF to generate the prediction. 


The above methods primarily focus on one-step prediction of public transport arrival time, which only consider the ATP of the next adjacent stop. The long-term multi-step prediction is also crucial for the public transport management system and it can provide more useful reference information for decisions of transport operators and travelers. With the development of deep neural networks (DNNs),  DNN-based time series prediction methods were  widely used in ATP.~\cite{Ratneswaran} proposed combining convolutional Long short-term memory (ConvLSTM) with ensemble learning and employing the eXtreme Gradient Boosting (XGBoost) statistical method to model heterogeneous traffic characteristics, thereby achieving prediction of bus arrival times under various traffic conditions.~\cite{liu2023understanding} developed a KF-LSTM (Kalman filter-LSTM) deep learning method to predict bus travel time. With the statistical analysis, it was found that the KF-LSTM approach can outperform the ensemble learning strategy.~\cite{he2020learning} proposed a traffic pattern-centric segment coalescing framework to learn heterogeneous traffic patterns and incorporated LSTM in each cluster to predict the bus travel time. Besides investigating the interaction of arrival time for public transport on a single route,~\cite{li2023sequence} designed a parallel Gated Recurrent Unit (GRU) network-based method to capture the spatial correlation among bus stops on different lines under the condition of limited data. In summary, with the advancement of data collection technologies (e.g., AVL data), deep learning-based methods demonstrate significant superiority in capturing nonlinear relationships, time series prediction, and multi-sources heterogeneous information fusion.


However, there are two inherent shortcomings for current solutions. Firstly, most methods for multi-step ATP focus on exploring the relationships between different time points alongside the one-dimensional temporal variation. Here, the one-dimensional variation refers to considering changes in features over a continuous duration. Specifically, for the public transport ATP, it involves analyzing the information at different stops on a specific route. While one-dimensional time series analysis helps capture continuity, periodicity and trends, real-world time series often have intricate temporal patterns (e.g., rising, falling, etc.). These different patterns are always mixed, overlapped with each other and their characteristics are obscured deeply in time series~\cite{wu2021autoformer,zhou2022fedformer,wu2022timesnet}. As shown in the bottom right of Fig.~\ref{fig_wait}, for a sequence of public transport arrival information with 39 stops, the temporal pattern of falling and fluctuation are hidden in the time series, which are labeled in red and green, respectively. These similar patterns are difficult to be captured by the one-dimensional based algorithms. Therefore, for the multi-step ATP, the modeling of multiple temporal variations is required to model the connection between similar temporal patterns. Secondly, most of the existing studies only used the data collected from one single or few routes~\cite{li2023transit,buchel2022we}, which are rather limited~\cite{ma2022multi}. It reduces the predictive capability for newly added, modified, or unmodeled routes. 


To address these issues, in this paper, a two-dimensional temporal variation-based bus/tram ATP model is proposed, which is termed  \textit{ArrivalNet}. It transforms one-dimensional time series into two-dimensional tensors to represent the intra-period and inter-period interactions. It can capture the implicit periodic information in the multi-step sequential prediction of arrival time. Meanwhile, a large scale city-wide dataset is constructed for validation~\cite{rong2022bus}. The primary contributions of this paper are summarized as follows:

\begin{itemize}
    \item A two-dimensional temporal variation-based bus/tram ATP model that can capture the intra-period and inter-period information in time series is proposed;
    \item The proposed model is validated on city-wide public transport data, which includes bus/tram operational data from 125 days. 
\end{itemize}

The remainder of this paper is organized as follows. The problem of public transport ATP is formulated in Section~\ref{section-problem-formulation}. Then, details of \textit{ArrivalNet} are presented in Section~\ref{method}. Section~\ref{experiments} shows the experimental setting and the comparison results. Finally, Section~\ref{conclusion} presents the conclusion and future work.

\begin{figure*}[t!]
  \centering
  \includegraphics[width=0.8\textwidth]{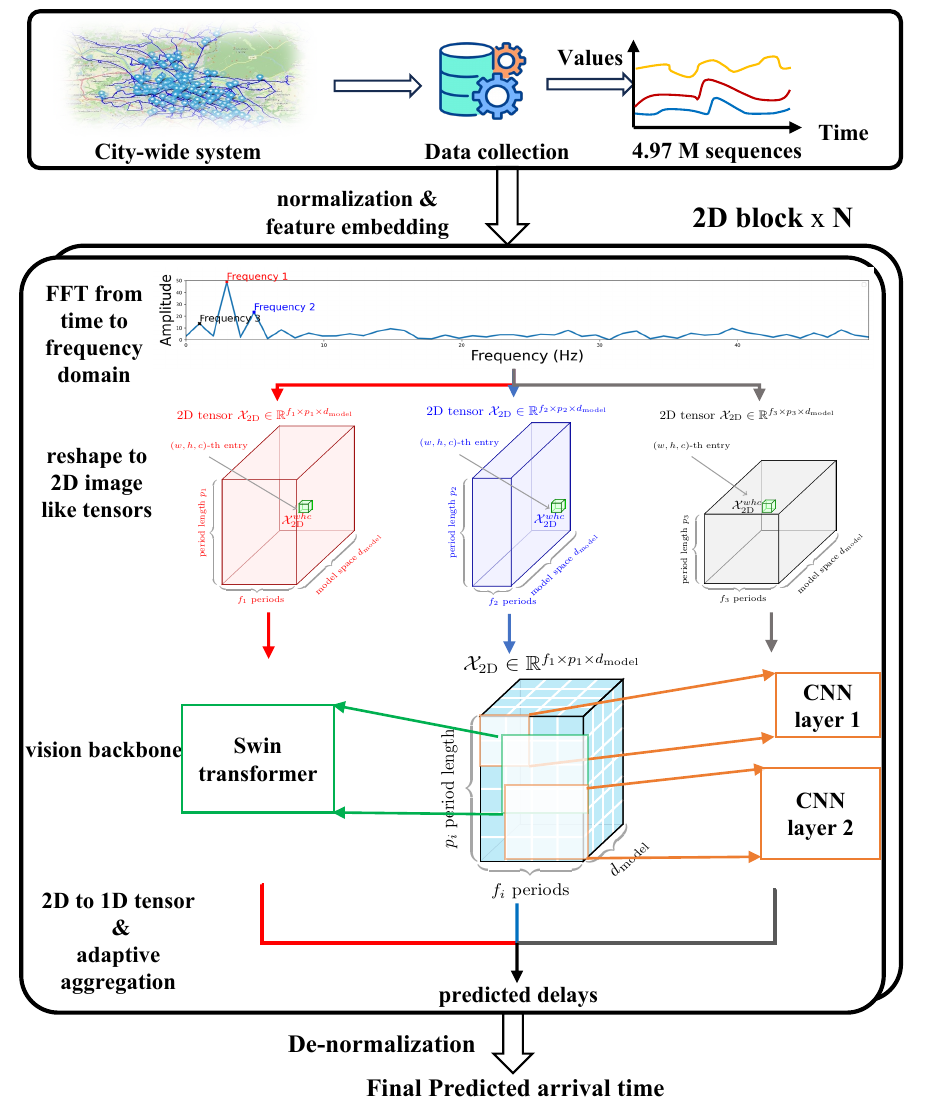}
  \caption{The overall framework of the proposed \textit{ArrivalNet}. It consists of five parts: feature embedding, fast fourier transform (FFT) from time to frequency domain,  reshape from 1D tensor to 2D image like tensor, vision backbone (CNN-based or Transformer-based) and adaptive aggregation. Note that two possible algorithms are shown, but only one is enough for feature extraction.}
  \label{fig_ArrivalNet}
\end{figure*}

\section{Problem formulation}\label{section-problem-formulation}
This paper focuses on the multi-step bus/tram ATP. The problem formulation of ATP is divided into three parts: the description of the ATP problem, the construction of variables, and the series stationarization. 
\subsection{The formulation of bus/tram arrival time prediction}
Following the formulation in~\cite{chen2023probabilistic}, in this work, the prediction of arrival time $T^{\text{a}}_{i}$ at stop $i$ is converted to the prediction of delay $T^{\text{d}}_{i}$ relative to the scheduled arrival time $T^{\text{s}}_{i}$, which implicitly includes the dwell time at stop $i-1$ in the travelling time between stops $i-1$ and $i$.
\begin{equation}\label{problem_convert}
    T^{\text{a}}_{i} = T^{\text{d}}_{i}+T^{\text{s}}_{i}
\end{equation}
With~(\ref{problem_convert}), the multi-step ATP is transformed into the sequential prediction of delay in future $N_{f}$ stops based on the information in the past $N_{p}$ stops, which can be formulated as follow:
\begin{equation}\label{prediction_eq}
\hat{\mathbf{T}}^{\text{d}}_{i+1:i+N_{f}} = f_{\mathbf{\theta}}(\mathbf{F}^{\text{temporal}}_{i-N_{p}+1:i},\mathbf{F}^{\text{static}})
\end{equation}
\begin{equation}\label{multi_step}
    \hat{\mathbf{T}}^{\text{a}}_{i+1:i+N_{f}} = \hat{\mathbf{T}}^{\text{d}}_{i+1:i+N_{f}}+\mathbf{T}^{\text{s}}_{i+1:i+N_{f}}
\end{equation}
where $\hat{\mathbf{T}}^{\text{a}}_{i+1:i+N_{f}}$, $\hat{\mathbf{T}}^{\text{d}}_{i+1:i+N_{f}}$ and $\mathbf{T}^{\text{s}}_{i+1:i+N_{f}}$ are predicted arrival time, predicted delay time and scheduled arrival time for stops ${i+1}$ to ${i+N_{f}}$, respectively. $\mathbf{F}^{\text{temporal}}_{i-N_{p}+1:i}$ are temporal features in the past $N_{p}$ stops and $\mathbf{F}^{\text{static}}$ is the static  contextual information, which can be flexibly designed and embedded as the input. $f_{\mathbf{\theta}}$ is the prediction model with trainable parameters $\mathbf{\theta}$.

\subsection{The construction of variables and series stationarization}
Considering that this work aims to develop a generic bus/tram ATP model, at each stop $i$, the index of the specific line is not considered as the input feature. At stop $i$, the temporal features $\mathbf{F}^{\text{temporal}}_{i-N_{p}+1:i}=[\mathbf{F}^{\text{}}_{i-N_{p}+1},...,\mathbf{F}^{\text{}}_i]^\top\in \mathbb{R}^{N_{p}\times C}$ are the serial combination of information $\mathbf{F}_{j}$ ($\forall j\in[i-N_{p}+1,i], \ \ j\in\mathbb{Z}$) in each past stop:

\begin{equation}
\mathbf{F}^{\text{}}_{j}=[S^{\text{t}}_{j},T^{\text{t}}_{j}, T^{\text{d}}_{j}, I_{j}, \overline{T}^{\text{t}}_{j}]
\end{equation}
where $C$ is the length of feature space. $S^{\text{t}}_{j}$ is the traveling distance between stop $j$ and previous stop $j-1$, which is highly related to the arrival time and can be obtained from the route of the specific line.  $T^{\text{t}}_{j}$ is the scheduled traveling time between stop $j$ and previous stop $j-1$ in the daily updated timetable. $T^{\text{d}}_{j}$ is the delay at stop $j$. It is calculated from the difference of timetable and real-time arrival information. $I_{j}$ is is a boolean value indicating whether there is a traffic light between stop $j$ and previous stop $j-1$. $\overline{T}^{\text{t}}_{j}$ is the average of $T^{\text{t}}_{j}$ at stop $j$ in collected data. $S^{\text{t}}_{j}$ and $T^{\text{t}}_{j}$ are daily updated from the centralized data collection system, which take into account of stop and route changing caused by adjustment in road infrastructure. The average travel time is calculated by statistically aggregating the time on the link between stop $j-1$ and $j$. A significant discrepancy between the actual travel time and the average travel time may indicate the presence of exceptional circumstances affecting the delay. 

\subsection{Series stationarization}
As for the data collected from the public transport system, the sequences always reflect non-stationarity, which is characterized by the continuous change of joint distribution over time. This reduces the predictability of time series. In~\cite{kim2021reversible}, an effective normalization-and-denormalization strategy was proposed to normalize instances with learnable parameters in the transformation from raw to normalized time series. It alleviated the temporal distributional shift by learning the affine transformation of input in the normalization and restoring the corresponding output in the denormalization. From the perspective of ATP, the shift of the temporal distribution is the variation of input features' joint distribution along the temporal dimension. For example, the corresponding relationship between average link travel time and distance may change along the same public transport line.~\cite{liu2022non} experimentally demonstrated that the algorithm also worked without learnable parameters. In normalization, it standardizes instances with varied means and variances. Then, the prediction is recovered with the same statistical properties in denormalization. 

For the original input sequence $\mathbf{F}^{\text{temporal}}_{i-N_{p}+1:i}$,  apply the normalization and de-normalization along the temporal dimension to obtain the normalized input ${\mathbf{F}^{\text{temporal}}_{i-N_{p}+1:i}}^{\prime}=[{\mathbf{F}^{\text{}}_{i-N_{p}+1}}^{\prime},...,{\mathbf{F}^{\text{}}_i}^{\prime}]^\top\in \mathbb{R}^{N_{p}\times C}$:
\begin{equation}
    \mu_{\text{}} = \frac{1}{N_{p}} \sum_{j=i-N_{p}+1}^{i} \mathbf{F}^{\text{}}_j
\end{equation}
\begin{equation}
\sigma_{\text{}}^2 = \frac{1}{N_{p}}\sum_{j=i-N_{p}+1}^{i} (\mathbf{F}^{\text{}}_j - \mu_{\text{}})^2
\end{equation}
\begin{equation}
{\mathbf{F}^{\text{}}_{j}}^{\prime} = \frac{1}{\sigma_{\text{}}} \odot (\mathbf{F}^{\text{}}_j - \mu_{\text{}}) \ \ \ \ 
 \forall j\in[i-N_{p}+1,i], \ \ j\in\mathbb{Z}
\end{equation}
where $\mu_{\text{}}$ and $\sigma_{\text{}}\in\mathbb{R}^{C\times 1}$ are mean and standard deviation of time series. $\odot$ is the element wise product. The normalized input feature ${\mathbf{F}^{\text{temporal}}_{i-N_{p}+1:i}}^{\prime}=[{\mathbf{F}^{\text{}}_{i-N_{p}+1}}^{\prime},...,{\mathbf{F}^{\text{}}_i}^{\prime}]^\top\in \mathbb{R}^{N_{p}\times C}$ is combined with $\mathbf{F}^{\text{static}}$ as  $\mathbf{F}_{\text{1D},i}\in \mathbb{R}^{N_{p}\times(C+N_{\text{c}})}$,  where $N_{\text{c}}$ is the feature dimension of contextual information. 
To simplify the expression, $\mathbf{F}_{\text{1D},i}$ is substituted with $\mathbf{F}_{\text{1D}}$. In the following parts, $\mathbf{F}_{\text{1D}}$ is the feature representation at a specific stop. Then, $\mathbf{F}_{\text{1D}}$ is sent to the prediction module to generate the estimation $\hat{\mathbf{T}}^{\text{d}\prime}_{i+1:i+N_{f}}=[\hat{\mathbf{T}}^{\text{d}\prime}_{i+1},...,\hat{\mathbf{T}}^{\text{d}\prime}_{i+N_{f}}]^\top\in \mathbb{R}^{N_{f}\times C}$.   Then, the de-normalization is operated for final multi-step predictions $\hat{\mathbf{T}}^{\text{d}}_{i+1:i+N_{f}}=[\hat{\mathbf{T}}^{\text{d}}_{i+1},...,\hat{\mathbf{T}}^{\text{d}}_{i+N_{f}}]^\top\in \mathbb{R}^{N_{f}\times C}$:
\begin{equation}
\hat{\mathbf{T}}^{\text{d}}_{k} = \sigma\odot\hat{\mathbf{T}}^{\text{d}\prime}_{k}+\mu_{\text{}} \ \ \ \ 
 \forall k\in[i+1,{i+N_{f}}], \ \ k\in\mathbb{Z}
\end{equation}
After the de-normalization, the outputs have the same feature space as the original inputs. The two-step operations can reduce the influence of non-stationarity in time series.

\section{Methodology}\label{method}


The proposed \textit{ArrivalNet} can be divided into five parts: feature embedding, fast fourier transform (FFT) from time to frequency domain,  reshape from 1D tensor to 2D image like tensor, vision backbone for feature extraction and adaptive aggregation, which are shown in Fig.~\ref{fig_ArrivalNet}. For the first part, the normalized input feature is expended from the original feture space to the model space, which increases the learning ability of model. The last four parts form the 2D block. It is the basic module of \textit{ArrivalNet}. In the FFT, the sequential input in the time domain is converted into frequency domain with different frequencies. Then, the 1D tensor is reshaped to various 2D tensors according to the frequencies and periods from FFT. The useful information in  two-dimensional image like features is captured by generic vision backbone (CNN and Swin Tranformer). the parameters from the periodic decomposition are designed to adaptively aggregate features of different periods.

\begin{figure*}[h]
  \centering
  \includegraphics[width=0.95\textwidth]{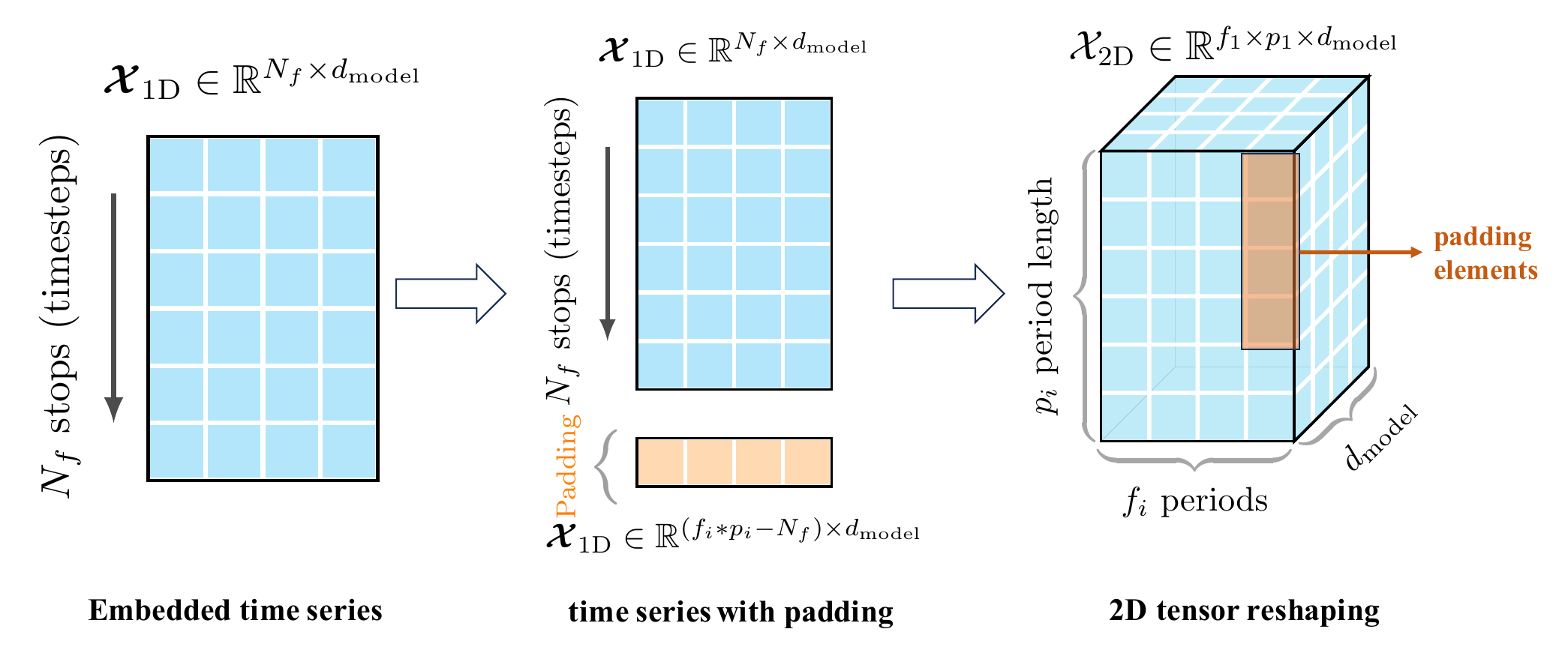}
  \caption{The illustration 2D tensor padding.}
  \label{fig_padding}
\end{figure*}

\subsection{Feature embedding}
The normalized feature ${\mathbf{F}^{\text{temporal}}_{i-N_{p}+1:i}}^{\prime}$ is combined with $\mathbf{F}^{\text{static}}$ as the final input sequence $\mathbf{F}_{\text{1D}}\in \mathbb{R}^{N_{p}\times(C+N_{\text{c}})}$.  Time series data is inherently sequential, with the order of $N_{p}$ data points carrying significant information about temporal dynamics. Unlike recurrent neural network (RNN)-based models that intrinsically understand sequence order, the two dimensional tensor of 2D block doesn't have a built-in mechanism to recognize the order of inputs. The positional encoder (\text{PE}) provide a way to incorporate this crucial information by adding a unique positional signal to each data point in the sequence, enabling the model to recognize and utilize the order of observations~\cite{vaswani2017attention}. Meanwhile, to  better understand and capture complex patterns in multivariate time series, including non-linear relationships and interactions between different features, a value encoder (\text{VE}) is applied to expand features to a high-dimensional model space  $d_{\text{model}}$ and increase model's learning ability~\cite{zhou2021informer}. Wtih $\mathbf{F}^{\text{static}}$ from the series stationarization, the formulation of \text{PE} and \text{VE} are shown as follow:
\begin{equation}
\mathbf{F}_{\text{PE}}^{(pos,2j)} = \sin\left({pos}/{(2N_{p})^{{2j}/{d_{\text{model}}}}}\right)
\end{equation}
\begin{equation}
\mathbf{F}_{\text{PE}}^{(pos,2j+1)} = \cos\left({pos}/{(2N_{p})^{{2j}/{d_{\text{model}}}}}\right)
\end{equation}
\begin{equation}
\mathbf{F}_{\text{VE}} = \underset{(C+N_{\text{c}})\rightarrow d_{\text{model}} }{\text{conv1D}}(\mathbf{F}_{\text{1D}})
\end{equation}
\begin{equation}\label{embed_linear}
\mathbf{{X}}_{\text{1D}}= \underset{N_{p}\rightarrow(N_{p}+N_{f}) }{\text{Linear}}(\mathbf{F}_{\text{PE}}+\mathbf{F}_{\text{VE}})
\end{equation}
 where $j\in \{1,...,\left\lfloor {d_{\text{model}}}/{2} \right\rfloor\}$ and ${pos}\in \{1,...,\left\lfloor N_{p} \right\rfloor\}$. $\mathbf{F}_{\text{PE}}\in\mathbb{R}^{N_{p}\times d_{\text{model}}}$ is the unique positional values. $\mathbf{F}_{\text{VE}}\in\mathbb{R}^{N_{p}\times d_{\text{model}}}$ is the encoded model features. $\text{conv1D}$ is the one dimension convolutional along the temporal dimension of $\mathbf{F}_{\text{1D}}$. It expend the feature space from lower to higher dimension. $\text{Linear}(\cdot)$ is the linear neural network to align the past sequence length $N_{p}$ to full sequence length $N_{p}+N_{f}$. In~(\ref{embed_linear}), the combination of $\mathbf{F}_{\text{PE}}$  and $\mathbf{F}_{\text{VE}}$ is the element-wise sum. To the end, $\mathbf{F}_{\text{1D}}$ is encoded as $\mathbf{X}_{\text{1D}}\in \mathbb{R}^{(N_{p}+N_{f})\times d_{\text{model}}}$.

\subsection{Fast fourier transform from time to frequency domain}\label{part-FFT}
It is well recognized that time series can be analyzed from two perspectives: the time domain and the frequency domain~\cite{koopmans1995spectral,box2015time}. The RNN-based and attention-based methods focus on the modeling of temporal relationships in the time domain, which is termed one-dimensional temporal variation~\cite{vaswani2017attention,hochreiter1997long}. The periodic trends (e.g., daily, weekly, monthly) can be reflected by the multi-periodic positional encoder~\cite{zhou2021informer}. However, some hidden periodic information may be neglected (e.g. rising, falling, fluctuation). It involves establishing connections between time points with the same time index across different periods. Fourier analysis serves as a common tool for transforming serial input from the time domain to the frequency domain~\cite{nussbaumer1982fast}.~\cite{zhou2022fedformer} designed a frequency-enhanced block to capture these obscured variations by Fast Fourier Transform (FFT). In~\cite{wu2022timesnet}, a similar strategy was used for the transformation from one-dimensional to two-dimensional tensors. In this work, FFT is selected as the tool to convert sequences from time to frequency domain. With the embedded feature $\mathbf{X}_{\text{1D}}$, the process of FFT is expressed as:
\begin{equation}
\mathbf{A} =  \text{FFT} \left( \mathbf{X}_{\text{1D}} \right) 
\end{equation}
\begin{equation}
\{ f_1, \cdots, f_{\frac{T}{2}} \} = \text{Avg} \left( \text{Amp} \left( \mathbf{A}\right) \right)
\end{equation}
where $\text{FFT}(\cdot)$ is the Fast Fourier Transform\footnote{FFT is implemented by Pytorch library: pytorch.org.} from the time domain to the frequency domain and $\mathbf{A}$ comprises a series of complex numbers, each representing the magnitude and phase of frequency components within the sequence. $\text{Amp}(\cdot)$ is the amplifying function of complex number and $\text{Avg}(\cdot)$ is the mean of elements in the feature dimension. $\{ f_1, \cdots, f_{\frac{T}{2}} \}$ is the obtained frequencies. Due to the conjugacy in the transformed  frequency domain, $f_{\ast}$ is only selected within $\{1, \cdots, \left\lfloor \frac{T}{2} \right\rfloor\}$.  Simply considering the effects of all frequencies will lead to a decrease in prediction performance because high-frequency changes in the sequence may be caused by noise. To avoiding the noise from meaningless high frequencies, only the most prominent $k$ frequencies are chosen.
\begin{equation}
\{ f_1, \cdots, f_k \} = \underset{f_{\ast} \in \{1, \cdots, \left\lfloor \frac{T}{2} \right\rfloor\}}{\text{arg Topk}}(\mathbf{A})
\end{equation}
\begin{equation}\label{fft}
p_i = \left\lfloor \frac{T}{f_i} \right\rfloor i \in \{1, \cdots, k\}
\end{equation}
where $\{ f_1, \cdots, f_k \}$ is the frequency of $k^{th}$ period with length $p_i$ and  the unnormalized calculated amplitude is as follows:
\begin{equation}\label{topk}
\mathbf{A}_{\text{Top K}} = \{\mathbf{A}_{f_{1}},...,\mathbf{A}_{f_{k}}\}
\end{equation}

\subsection{Reshape from 1D to 2D image like tensor}\label{part-reshape}
Based on (\ref{fft}) and (\ref{topk}), for each frequency $f_{j}, j\in[1,k] \ \ j\in\mathbb{Z}$, the one-dimensional time serie $\mathbf{X}_{\text{1D}}\in \mathbb{R}^{(N_{p}+N_{f})\times d_{\text{model}}}$ can be transformed and spliced into a tensor $\mathbf{X}^{j}_{\text{2D}}\in \mathbb{R}^{f_{j}\times p_{j}\times d_{\text{model}}}$. $\text{2D}$ means that the inter-periodic and intra-periodic information is represented in $\mathbf{X}^{j}_{\text{2D}}$. It is an image-like tensor with the channel length $d_{\text{model}}$. Each row of the tensor contains the time points within a period, and each column of it involves the time points at the same intra-periodic index  across various periods. Ideally, the relationship of $f_{j}$, $p_{j}$, $j\in[1,k] \ \ j\in\mathbb{Z}$ is $f_{j}\times p_{j} = (N_{p}+N_{f})$. However, in some cases,~(\ref{fft}) cannot be divisible integrally. A padding operation is needed for $\mathbf{X}_{\text{1D}}$.
\begin{equation}
\mathbf{X}^{\text{padding}}_{\text{1D}} = \underset{p_j, f_j}{ \text{Padding}}(\mathbf{X}_{\text{1D}}), \quad j \in \{1, \ldots, k\}
\end{equation}
\begin{equation}
\mathbf{X}^{j}_{\text{2D}} = \text{Reshape}_{p_j, f_j} \left( \mathbf{X}^{\text{padding}}_{\text{1D}} \right), \quad j \in \{1, \ldots, k\}
\end{equation}
where $\text{Padding}(\cdot)$ is applied to impute the feature with zeros for successful reshape. Fig.~\ref{fig_padding} illustrates the process of padding and reshape.

\begin{figure*}[h]
  \centering
  \includegraphics[width=0.95\textwidth]{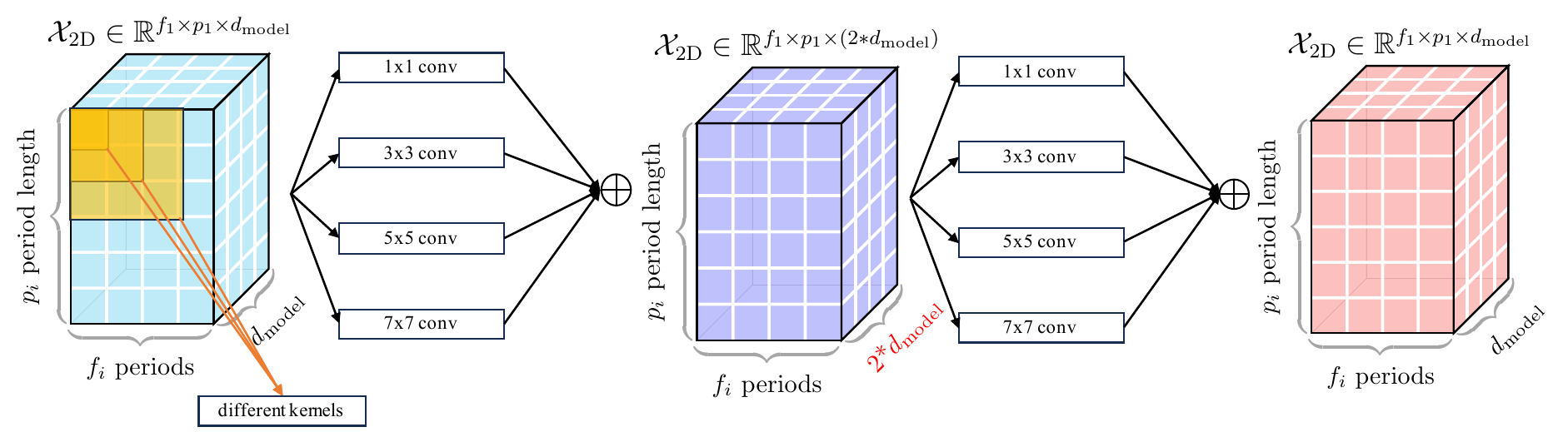}
  \caption{CNN-based feature extraction.}
  \label{fig_CNN}
\end{figure*}

\begin{figure*}[h]
  \centering
  \includegraphics[width=0.95\textwidth]{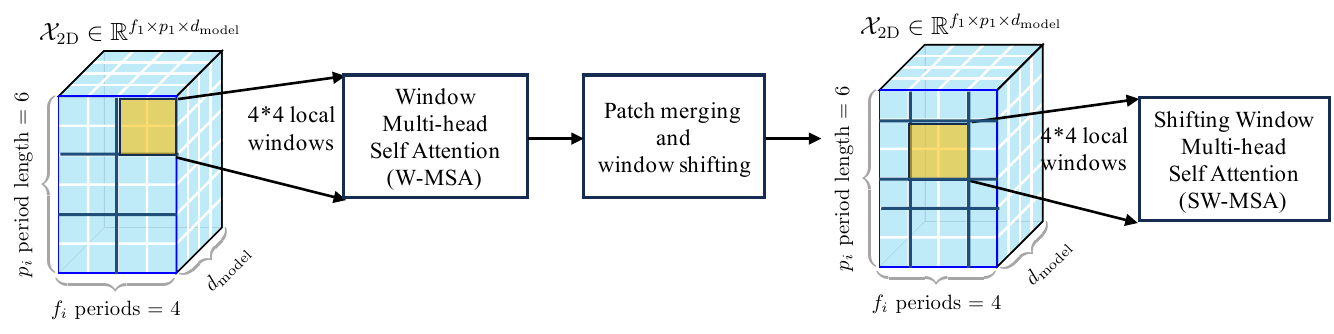}
  \caption{Swin Transformer-based feature extraction.}
  \label{fig_swin}
\end{figure*}

\subsection{Vision backbone}
Two popular deep-learning architectures, CNN and Transformer, are chosen for extracting features from the 2D tensors. We remark that the proposed prediction model is generic and allows the use of other architectures. 
\subsubsection{CNN-based feature extraction}
For each element $\mathbf{X}^{j}_{\text{2D}}\quad j \in \{1, \ldots, k\}$ in $\mathbf{X}_{\text{2D}}$, the first and second dimensions represent the intra-periodic and inter-periodic information, which is similar to the feature space of image in computer vision. The frequency $f$, period $p$ and $d_{\text{model}}$ dimensions are corresponding to the height, width and channel dimensions in an image. Therefore, the tensor $\mathbf{X}_{\text{2D}}$ can be easily processed by the parameter-efficient inception block and finally transformed back to one-dimensional feature space~\cite{szegedy2015going}:
\begin{equation}
\hat{\mathbf{X}}^{j}_{\text{2D}}=\text{Inception}_{\text{Conv2d}}(\mathbf{X}^{j}_{\text{2D}}), \quad j \in \{1, \ldots, k\}
\end{equation}
\begin{equation}\label{2d-1d}
\hat{\mathbf{X}}^{j}_{\text{1D}}=\text{Reshape}(\hat{\mathbf{X}}^{j}_{\text{2D}}), \quad j \in \{1, \ldots, k\}
\end{equation}
where $\hat{\mathbf{X}}^{j}_{\text{2D}}\in \mathbb{R}^{f_{j}\times p_{j}\times d_{\text{model}}}$ is the tensor processed by $\text{Inception}(\cdot)$ and $\hat{\mathbf{X}}^{j}_{\text{1D}}\in \mathbb{R}^{(N_{p}+N_{f})\times d_{\text{model}}}$ is the output feature of the module. As shown in Fig.~\ref{fig_CNN}, in the parameter-efficient inception block, kernels with different sizes operate the two dimensional convolution along dimensions $f*p$. All kernels work parallelly and are summed together in the first layer. Then, with a nonlinear actvation function, it is sent to the second layer with the similar operation. By carefully designing the relationship of kernel size and padding length in each layer, the input and output of $\text{Inception}(\cdot)$ are tensors with same size, which doesn't influence the transformation from two to one dimensional tensor.

From the analysis in Section~\ref{part-FFT} and~\ref{part-reshape}, the 1-dimensional tensor obtained by feature embedding is transformed into 2D tensor after FFT and padding processes. For the designed information extractor within the 2D block, besides employing a CNN-based inception model to extract useful features from the image-like tensor, other popular vision backbones can also be utilized. In this work, an attention-based approach, named Swin Transformer, is applied to the 2D tensor.

\begin{figure}[h]
  \centering
  \includegraphics[width=0.48\textwidth]{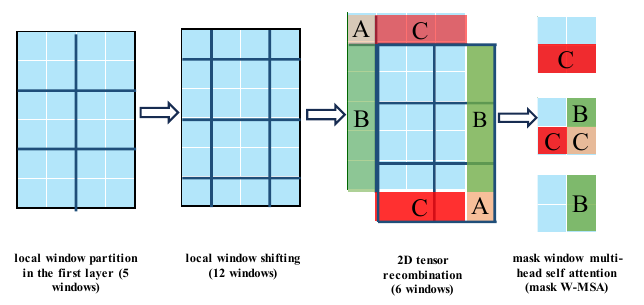}
  \caption{The computationally efficient operation after the window shifting.}
  \label{fig_swin_batch}
\end{figure}

\subsubsection{Attention-based feature extraction}
In deep learning, the development of Transformer has significantly enhanced model performance in sequential analysis and natural language processing (NLP) based on the attention mechanism. However, for vision tasks, transformer-based models struggle to process image information efficiently. The difficulty arises from the high resolution of the image, while calculating attention across all pixels is computationally prohibitive. To address this issue,~\cite{dosovitskiy2020image} proposed dividing the image into several non-overlapping patches and extracting useful information through a hierarchical structure. However, compared to CNN-based methods, this approach struggles to establish connections between adjacent pixels across different patches, and the computation time increases quadratically with image size. Inspired by the sliding operation in CNNs,~\cite{liu2021swin} introduced a local transformer method based on sliding windows, termed Swin Transformer. It models adjacent pixels through the movement of local windows, while its computational complexity increases linearly relative to the image size. Therefore, in this work, the Swin Transformer is employed to extract useful information from image-like 2D tensors. 

As illustrated in Fig.~\ref{fig_swin}, we assume that the frequency and period length of the 2D tensor are 4 and 6, respectively. Compared to high-resolution images, the number of pixels in a 2D tensor is relatively low. Hence, the patch size is set to 1 and each 2x2 local window contains only 4 pixels, and the 2D tensor comprises 6 windows. After applying Window-based Multi-head Self Attention (W-MSA) within local windows, features from different local windows are recombined. The self-attention  mechenism is formulated as follow:
\begin{equation}
    \text{Attention}(\mathbf{Q}, \mathbf{K}, \mathbf{V}) = \text{Softmax}\left(\frac{\mathbf{Q}\mathbf{K}^{\top}}{\sqrt{ d_{\text{model}}}}\right)\mathbf{V}
\end{equation}
where $\mathbf{Q}, \mathbf{K}, \mathbf{V}\in \mathbb{R}^{M^{2}\times d_{\text{model}}}$ are \textit{query}, \textit{key} and \textit{value}, respectively. $M^2$ is the number of pixels (patches) in a local window. In the self-attention, $\mathbf{Q}$, $\mathbf{K}$, $\mathbf{V}$ are same features. To introduce cross-window connections while maintaining efficient computation of non-overlapping windows, a shifted window partitioning strategy is implemented. This involves moving the windows in the 2D tensor starting from the top left by the half size of a local window. In this case, it will expand the number of local windows from 6 to 12. To reduce the computational complexity,~\cite{liu2021swin} proposed an efficient computation approach based on the mask operation in the attention mechanism, which is detailed in Fig.~\ref{fig_swin_batch}. It shifts some partited windows to the opposite positions and make a re-partition of 2D tensor. To aovid unreasonable connection of pixels in the original 2D tensor, the mask matrices are generated to block the relationship between non-connected pixels. This efficient operation ensures that the computational complexity remains consistent with that of the initial window partition. The masked attention mechenism is formulated as follow:
\begin{equation}
    \text{Mask Att}(\mathbf{Q}, \mathbf{K}, \mathbf{V}, \mathbf{M}) = \text{Softmax}\left(\frac{\mathbf{Q}\mathbf{K}^{\top}}{\sqrt{d_{\text{model}}}} + \mathbf{M}\right)\mathbf{V}
\end{equation}
where $\mathbf{M}\in \mathbb{R}^{M^{2}\times M^{2}}$ represents the mask matrix that is added to the scores resulting from $\mathbf{Q}\mathbf{K}^{\top}$. The masked element is set to a large negative value. The shifting process and mask attention mechenism are collectively referred to as Shifting Window multi-head Self Attention (SW-MSA). Based on W-MSA, SW-MSA and the 2D tensor input $\mathbf{X}^{j}_{\text{2D}}\quad j \in \{1, \ldots, k\}$, the entire Swin Transformer consists of two layers and is computed as:
\begin{equation}
    \hat{\mathbf{X}}^{j,1}_{\text{2D}} = \text{W-MSA}(\text{LN}(\mathbf{X}^{j}_{\text{2D}})) + \mathbf{X}^{j}_{\text{2D}}
\end{equation}
\begin{equation}
    \mathbf{X}^{j,1}_{\text{2D}} = \text{MLP}(\text{LN}(\hat{\mathbf{X}}^{j,1}_{\text{2D}})) + \hat{\mathbf{X}}^{j,1}_{\text{2D}}
\end{equation}
\begin{equation}
    \hat{\mathbf{X}}^{j,2}_{\text{2D}} = \text{SW-MSA}(\text{LN}(\mathbf{X}^{j,1}_{\text{2D}})) + \mathbf{X}^{j,1}_{\text{2D}}
\end{equation}
\begin{equation}
     \mathbf{X}^{j,2}_{\text{2D}} = \text{MLP}(\text{LN}(\hat{\mathbf{X}}^{j,2}_{\text{2D}})) + \hat{\mathbf{X}}^{j,2}_{\text{2D}}
\end{equation}
where $\text{LN}(\cdot)$ and $\text{MLP}(\cdot)$ are layernorm and multi-layer peception, respectively. $\hat{\mathbf{X}}^{j,1}_{\text{2D}}$ and $\hat{\mathbf{X}}^{j,2}_{\text{2D}}$ are the output of local window attention. Similar to~(\ref{2d-1d}), the output of second layer $\mathbf{X}^{j,2}_{\text{2D}}$ can be transformed back to one-dimensional tensor.

\subsection{Adaptive output aggregation}
\begin{figure}[t]
  \centering
  \includegraphics[width=0.38\textwidth]{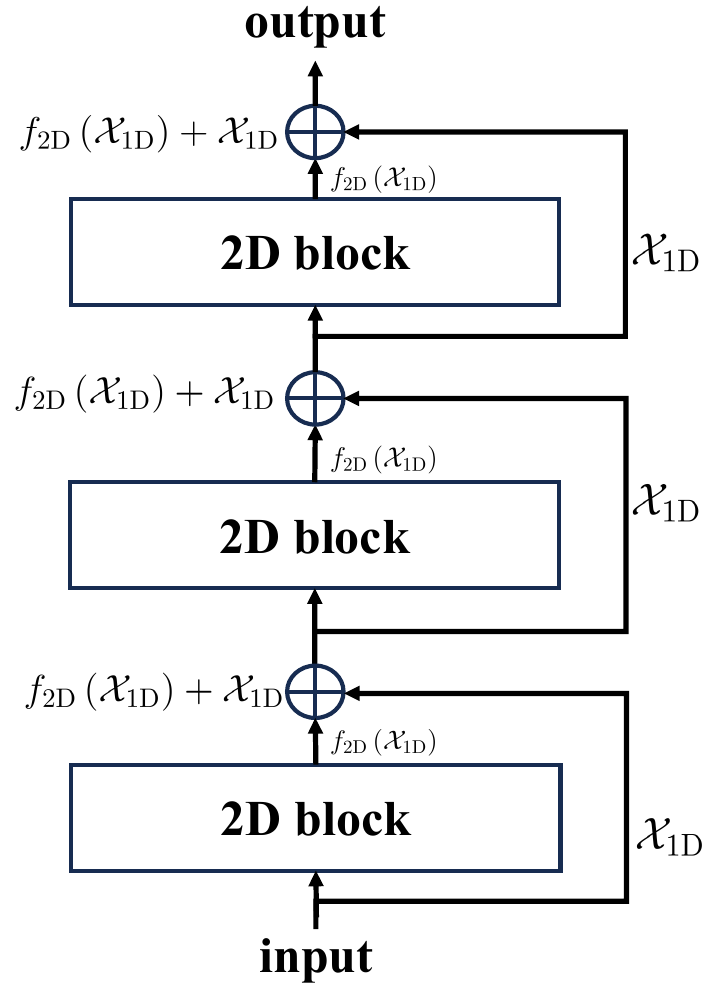}
  \caption{The residual connection of 2D blocks.}
  \label{fig_resnet}
\end{figure}
Based on the useful feature extraction of vision backbone, the formulation of multi-step ATP is converted into the feature extraction from the image-like tensor. Inspired by the well-developed theory of residual connection between different layers~\cite{he2016deep}, the developed 2D block, are stacked and connected by the residual learning framework. If the  \textit{ArrivalNet} consists of $L$ layers, for the $l^{th}\quad l \in \{1, \ldots, L\}$ layer, the residual connection is formulated as:
\begin{equation}
\mathbf{X}^{l}_{\text{1D}} = f_{\text{2D}} \left(\mathbf{X}^{l-1}_{\text{1D}} \right) + \mathbf{X}^{l-1}_{\text{1D}}
\end{equation}
where $f_{\text{2D}}(\cdot)$ is the 2D block. The residual connection is illustrated in Fig.~\ref{fig_resnet}. The processed outputs $\mathbf{X}^{j}_{\text{2D}}\quad j \in \{1, \ldots, k\}$ need to be aggregated  adaptively based on the normalized weight $\mathbf{A}_{\text{Top K}} = \{\mathbf{A}_{f_{1}},...,\mathbf{A}_{f_{k}}\}$ in periodic decomposition.
\begin{equation}
\hat{\mathbf{A}}_{\text{Top K}} =\text{Softmax}(\mathbf{A}_{\text{Top K}})
\end{equation}
\begin{equation}
\mathbf{X}_{\text{1D}} = \sum_{j=1}^{k} \mathbf{A}_{f_{j}} \times \hat{\mathbf{X}}_{\text{1D}}
\end{equation}
where $\text{Softmax}(\cdot)$ is the normalization of weight. $\mathbf{X}_{\text{1D}}\in \mathbb{R}^{(N_{p}+N_{f})\times d_{\text{model}}}$ is the output of \text{2D Block}. And the estimation $\hat{\mathbf{T}}^{\text{delay}}_{i+1:i+N_{f}}$ of the proposed \textit{ArrivalNet} is obtained by aligning $d_{\text{model}}$ back to one and cut the last $N_{f}$ elements in the temporal dimension, which means only predicting delays. 
\begin{equation}
\hat{\mathbf{T}}^{\text{delay}}_{i+1:i+N_{f}} = \text{Trun}(\underset{d_{\text{model}}\rightarrow 1}{\text{Linear}}(\mathbf{X}_{\text{1D}}))
\end{equation}
where $\text{Linear}(\cdot)$ is the linear neural network to embed the feature space from $d_{\text{model}}$ to 1. $\text{Trun}(\cdot)$ is the truncation for estimated $\hat{\mathbf{T}}^{\text{delay}}_{i+1:i+N_{f}}$.

\section{Experiments}\label{experiments}
To validate the performance of \textit{ArrivalNet} for bus/tram multi-step ATP, 125 days of public transport operational data in Dresden, Germany is collected. In this section, we describe the dataset, experimental settings (including evaluation metrics and comparison baselines), and results. 

\begin{figure}[t!]
  \centering
  \begin{subfigure}[b]{0.48\textwidth}
    \includegraphics[width=1\textwidth]{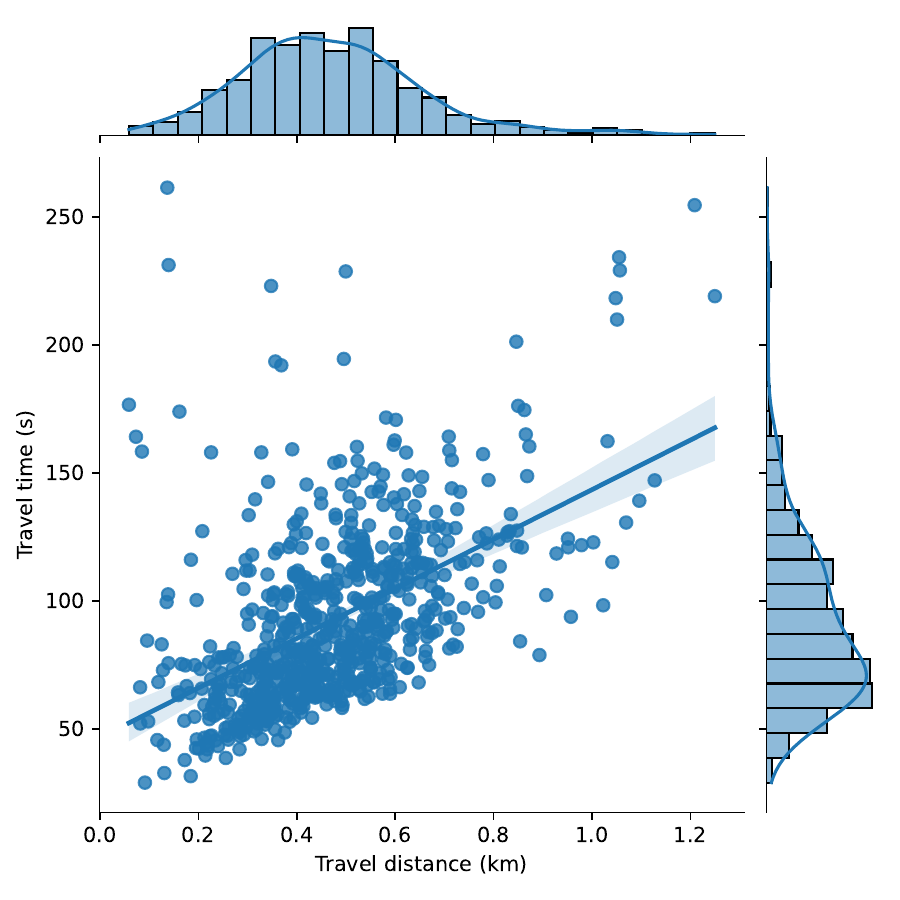}
    \caption{The joint distribution of tram. The average values of link travel distance and link travel time are 0.4702 km and 92.19 s, respectively.}
    \label{fig_relation_distance_and_time_1}
  \end{subfigure}
  
  \begin{subfigure}[b]{0.48\textwidth}
    \includegraphics[width=1\textwidth]{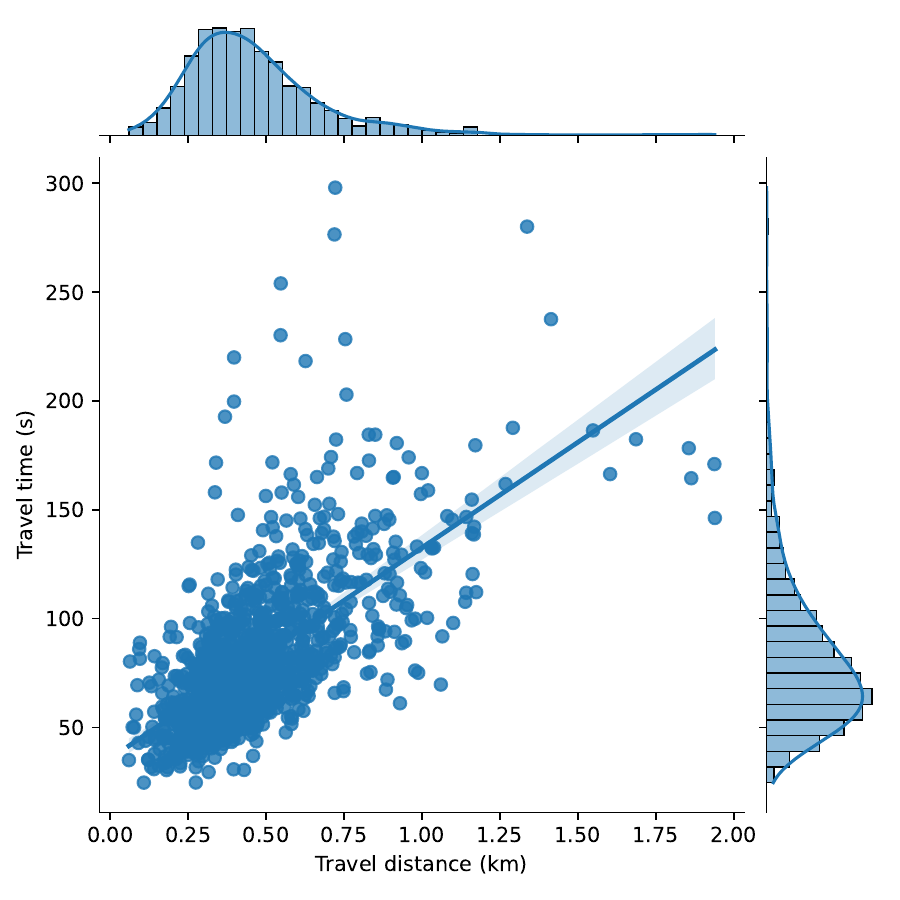}
    \caption{The joint distribution of tram. The average values of link travel distance and link travel time are 0.46 km and 80.79 s, respectively.}
    \label{fig_relation_distance_and_time_2}
  \end{subfigure}
  \caption{The relationship between link travel distance and link travel time.}
    \label{fig_relation_distance_and_time}
\end{figure}

\subsection{Dataset}
\subsubsection{Basic description of DVB data}
The urban public transport system in Dresden, Germany, is operated by DVB (Dresdner Verkehrsbetriebe AG) and consists of two modes: tram and bus. All operating public transport vehicles send their real-time status to the central data collection system at approximately 15 second intervals, which includes current time, current location, and distance from the previous stop within the route segment. In this work, only the status for the time of arrival is used. Additionally, the public transport system collects daily updated information on operating routes and stop locations, which may vary due to factors like road construction. In the constructed dataset, operational tram/bus data in 125 days from all bus and tram routes is collected. 4.97M valid sequences were extracted, where each sequence is the  sequential status of a bus/tram at the stop from departure platform to the destination platform. The massive and useless information for the bus/tram on the link between platforms is removed. The average values of link travel distance and link travel time for tram are 0.4702 km and 92.19 s, respectively. For bus, these average characteristics are 0.46 km and 80.79 s. In Fig.~\ref{fig_relation_distance_and_time}, the relationship of link travel distance and link travel time for tram and bus is presented. It indicates a positive correlation between travel distance and travel time, where an increase in travel distance tends to lead to an increase in travel time. This, in turn, impacts the ATP. Furthermore, it demonstrates the rationality of using travel distance and time as one of the selected features in Section~\ref{section-problem-formulation}.

\begin{figure}[t!]
  \centering
  \begin{subfigure}[b]{0.48\textwidth}
    \includegraphics[width=\textwidth]{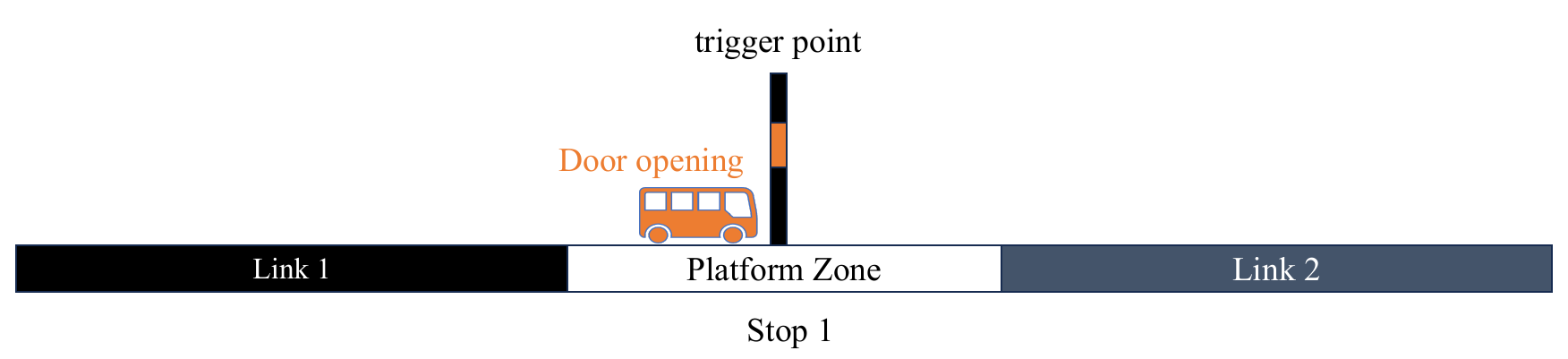}
    \subcaption{The vehicle does not pass the trigger point between the current and next stop and opens the door.}
    \label{fig:sub1}
  \end{subfigure}%

  \begin{subfigure}[b]{0.48\textwidth}
    \includegraphics[width=\textwidth]{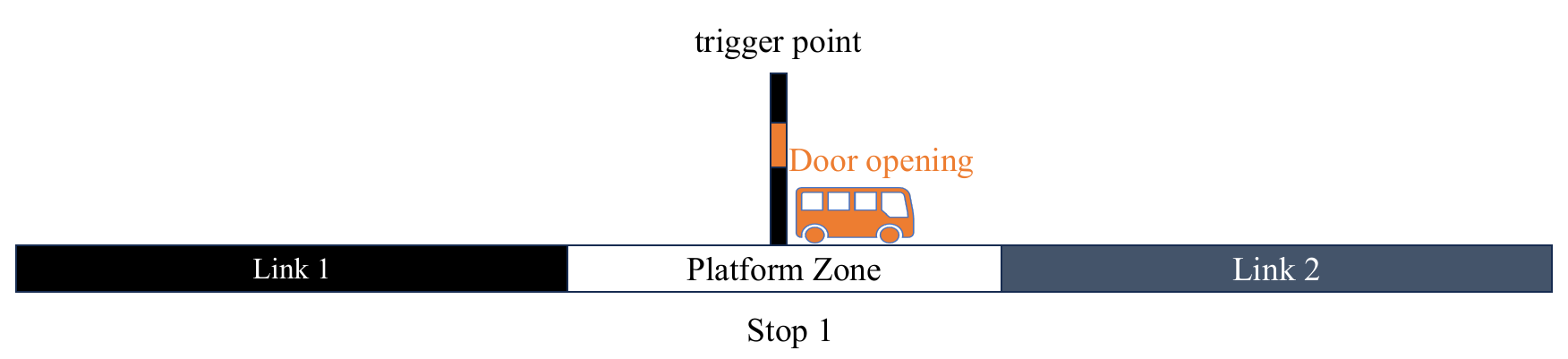}
    \subcaption{The vehicle passes the trigger point between the current and next stop and opens the door.}
    \label{fig:sub2}
  \end{subfigure}%
  
  \begin{subfigure}[b]{0.48\textwidth}
    \includegraphics[width=\textwidth]{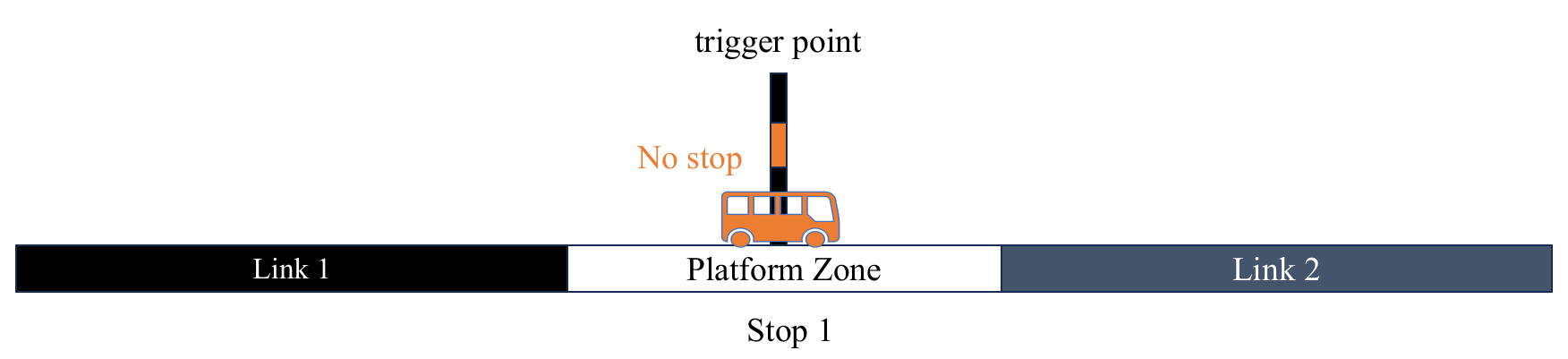}
    \subcaption{The vehicle passes the stop without opening doors.}
    \label{fig:sub3}
  \end{subfigure}
  \caption{Three conditions when the vehicle arrives at the stop.}
  \label{fig_capture_delay}
\end{figure}

\begin{figure}[t]
  \centering
  \includegraphics[width=0.48\textwidth]{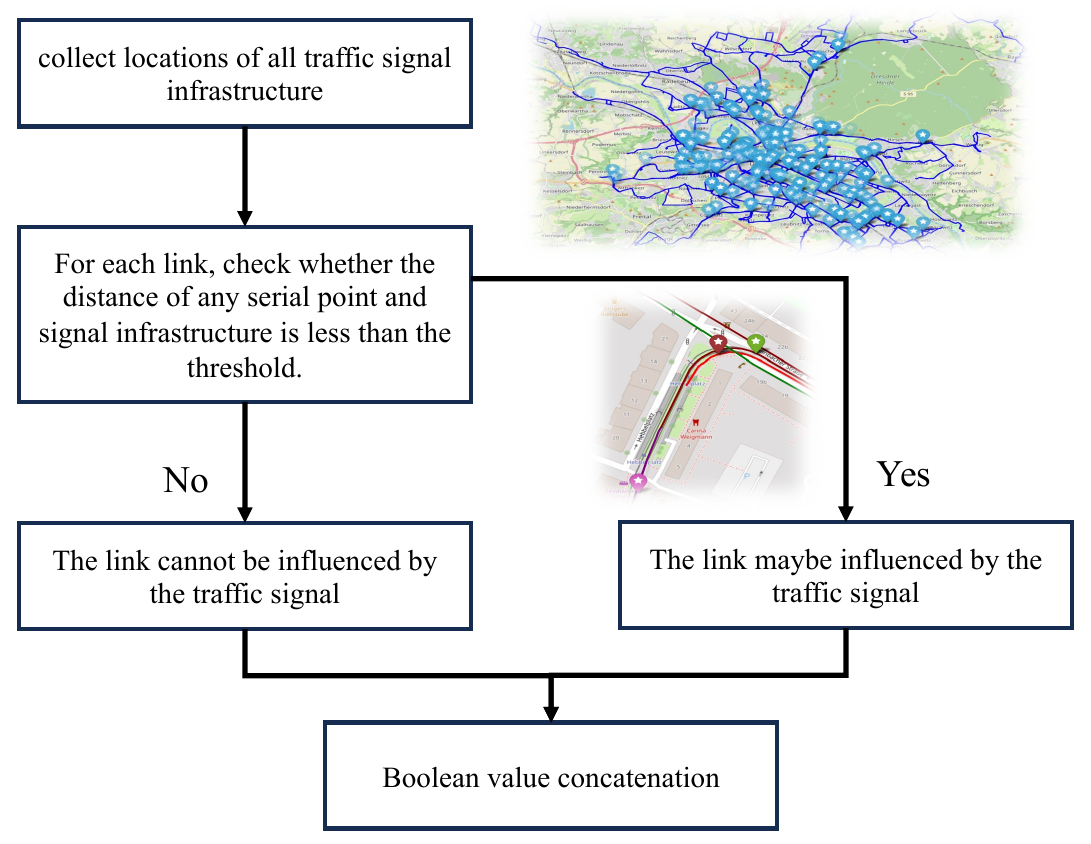}
  \caption{The selecting process of related traffic signal infrastructure.}
  \label{fig_capture_map}
\end{figure}

\subsubsection{The calculation of delays}
In this work, the ATP is formulated as predicting the delay time at each future stop in (\ref{multi_step}) for a specific vehicle. In the data pre-processing, it is necessary to accurately capture the actual arrival time to calculate the real delay in seconds. For the collected data, the enabling information of the door opening operation can serve as an indicator of vehicle arrival time, which is event-triggered and not affected by the approximate 15-second transmission frequency. Specifically, vehicle arrivals at the stop can be categorized into three conditions: the vehicle does not pass the trigger point between the current and next stop \& opens the door, the vehicle passes the trigger point between the current and next stop \& opens the door, and the vehicle passes the stop without opening doors. The details and differences are illustrated in Fig.~\ref{fig_capture_delay}. With these three conditions, the arrival time is precisely captured, which is applied for the calculation of delay in~(\ref{multi_step}).
\subsubsection{The selection of contextual information}
In this work, contextual information is incorporated into the \textit{ArrivalNet}, which can improve the performance of the model. Whether a link connecting two stops passes through a traffic signal infrastructure is the dynamic contextual feature, which may change temporally in a sequence. The matching process between infrastructure and daily updated public transport routes in Dresden is illustrated in Fig.~\ref{fig_capture_map}. The criterion for judgment is whether the distance between the traffic signal infrastructure's position and any waypoint in the link is less than a threshold. $\mathbf{F}^{\text{static}}$ consists of two static and contextual binary features: peak/off peak hour and weekday/weekend. The morning rush hour is set from 7:00 AM to 9:00 AM, and the evening rush hour is set from 4:00 PM to 7:00 PM on weekdays. The selection of related traffic signal infrastructure will be detailed in the experiments section.

\begin{table}[htbp]
\captionsetup{font={small}}
\caption{The setting of parameters in \textit{ArrivalNet}}
\label{table_parameters}
\centering
\begin{tabular}{ccc}
\toprule
\textbf{Description} & \textbf{Notation} & \textbf{Value} \\
\midrule
learning rate & - & 0.001 \\
embedding feature length & $d_{\text{model}}$ &16 \\
number of 2D block & - & 2 \\
selected frequency & $k$ & 3 \\
number of kernels in CNN & - & 6 \\
input series length & $N_{p}$ & 10 \\
output series length & $N_{f}$ & 5, 10 \\
local window size & - & 2 \\
threshold in related signal selection (m)& - & 20 \\
\bottomrule
\end{tabular}
\end{table}

\begin{table*}[h!]
  \begin{center}
   \captionsetup{font={small}}
    \caption{The Comparative Results of Tram arrival Time Prediction. The average of each metric is the mean of 10$\rightarrow$5 and 10$\rightarrow$10.}
    \label{table_tram}
    \begin{tabular}{
            >{\centering\arraybackslash}p{1.5cm}
            >{\centering\arraybackslash}p{1.5cm}
            >{\centering\arraybackslash}p{1.5cm}
            >{\centering\arraybackslash}p{1.5cm}
            >{\centering\arraybackslash}p{1.5cm}
            >{\centering\arraybackslash}p{1.5cm}
            >{\centering\arraybackslash}p{1.5cm}
            >{\centering\arraybackslash}p{1.5cm}
            >{\centering\arraybackslash}p{1.5cm}} 
    \toprule
      \textbf{Metric}  & \textbf{Length} & \textbf{ARIMA} & \textbf{LSTM} & \textbf{Transformer} & \textbf{TCN} & \textbf{\textit{ArrivalNet-1} (wo context)} & \textbf{\textit{ArrivalNet-2} (Swin)} & \textbf{\textit{ArrivalNet-3} (CNN)} \\
      \midrule
       \multirow{2}{*}{\parbox{1.5cm}{\centering RMSE \\(second)}}
            & 10$\rightarrow 5$ & 64.3 & 60.4 & 49.8 & 49.2 & 47.9 &\underline{46.4} & \textbf{46.2} \\
            & 10$\rightarrow 10$ & 79.3 & 72.4 & 69.1 & 68.5 & 61.4 &\textbf{56.8} & \underline{57.4} \\
            & Average & 71.8 & 66.4 & 59.45 & 58.85 & 54.65 &\textbf{51.6} & \underline{51.8} \\
      \hline
      \multirow{2}{*}{\parbox{1.5cm}{\centering MAE \\(second)}}
             & 10$\rightarrow 5$ & 55.2 & 49.9 & 38.8 & 42.5& 34.2 & \underline{32.7}  & \textbf{31.5} \\
            & 10$\rightarrow 10$ & 61.2 & 57.7 & 50.7 & 48.6 & 41.1 &\underline{38.6} & \textbf{37.4} \\
            & Average &            58.2 & 53.8  &44.75 &45.55 &37.65 &\underline{35.65} &   \textbf{34.45} \\
    \hline
          \multirow{2}{*}{\parbox{1.5cm}{\centering MAPE (\%)}}
         & 10$\rightarrow 5$ & 4.94 & 4.86 & 3.83 &3.59 & 2.73&\underline{2.54} & \textbf{2.39} \\
        & 10$\rightarrow 10$ & 6.71 & 5.12 & 4.32 &3.93 & 3.15&\underline{2.66} & \textbf{2.55} \\
        & Average &            5.825& 4.99 & 4.075& 3.76 & 2.94 & \underline{2.6}  & \textbf{2.47}\\
      \bottomrule
    \end{tabular}
  \end{center}
  \begin{tablenotes}
\item The $1^{st}$/$2^{nd}$ best results are indicated in \textbf{bold}/\underline{underline}.
\end{tablenotes}
\end{table*}

\begin{table*}[h!]
  \begin{center}
  \captionsetup{font={small}}
    \caption{The Comparative Results of Bus arrival Time Prediction. The average of each metric is the mean of 10$\rightarrow$5 and 10$\rightarrow$10.}
    \vspace{2ex}
    \label{table_bus}
    \begin{tabular}{
            >{\centering\arraybackslash}p{1.5cm}
            >{\centering\arraybackslash}p{1.5cm}
            >{\centering\arraybackslash}p{1.5cm}
            >{\centering\arraybackslash}p{1.5cm}
            >{\centering\arraybackslash}p{1.5cm}
            >{\centering\arraybackslash}p{1.5cm}
            >{\centering\arraybackslash}p{1.5cm}
            >{\centering\arraybackslash}p{1.5cm}
            >{\centering\arraybackslash}p{1.5cm}} 
    \toprule
      \textbf{Metric}  & \textbf{Length} & \textbf{ARIMA} & \textbf{LSTM} & \textbf{Transformer} & \textbf{TCN} & \textbf{\textit{ArrivalNet-1} (wo context)} & \textbf{\textit{ArrivalNet-2} (Swin)} & \textbf{\textit{ArrivalNet-3} (CNN)} \\
      \midrule
      \multirow{2}{*}{\parbox{1.5cm}{\centering RMSE \\(second)}}
 & 10$\rightarrow$5 & 66.1 & 63.7 & 53.1 & 52.8 & 49.3 & \underline{48.7} & \textbf{48.3}\\ 
 & 10$\rightarrow$10 & 83.3 & 74.2 & 65.1 & 67.9 &  61.9 & \textbf{57.3} & \underline{58.3}\\
 & Average &        74.7 & 68.95& 59.1 & 60.35& 55.6 & \textbf{53}  & \underline{53.3}\\
      \hline
      \multirow{2}{*}{\parbox{1.5cm}{\centering MAE \\(second)}} 
& 10$\rightarrow$5 & 53.8 & 49.5 & 39.2 & 41.4 & 36.1 & \underline{34.2} & \textbf{33.4}\\ 
& 10$\rightarrow$10 & 61.4 & 59.3 & 51.0 & 49.8 & 42.7 & \underline{38.5}& \textbf{38.3}\\
& Average &        57.6 &  54.4 &  45.1 &  45.6 &  39.4 &  \underline{36.35} &  \textbf{35.85}\\   
      \hline
      \multirow{2}{*}{\parbox{1.5cm}{\centering MAPE (\%)}}    
& 10$\rightarrow$5 & 5.84 & 5.32 & 3.94 & 3.91 & 3.67 & \underline{3.42}& \textbf{3.35}\\ 
& 10$\rightarrow$10 & 8.73 & 6.30 & 4.27 & 4.39 & 3.87 & \textbf{3.65}& \underline{3.74}\\
& Average &         7.285& 5.81 & 4.105& 4.15 & 3.77 & \textbf{3.535}& \underline{3.545}\\
      \bottomrule
    \end{tabular}
  \end{center}
  \begin{tablenotes}
\item The $1^{st}$/$2^{nd}$ best results are indicated in \textbf{bold}/\underline{underline}.
\end{tablenotes}
\end{table*}

\begin{figure*}[h!]
  \centering
  \begin{subfigure}[b]{0.48\textwidth}
    \includegraphics[width=1\textwidth]{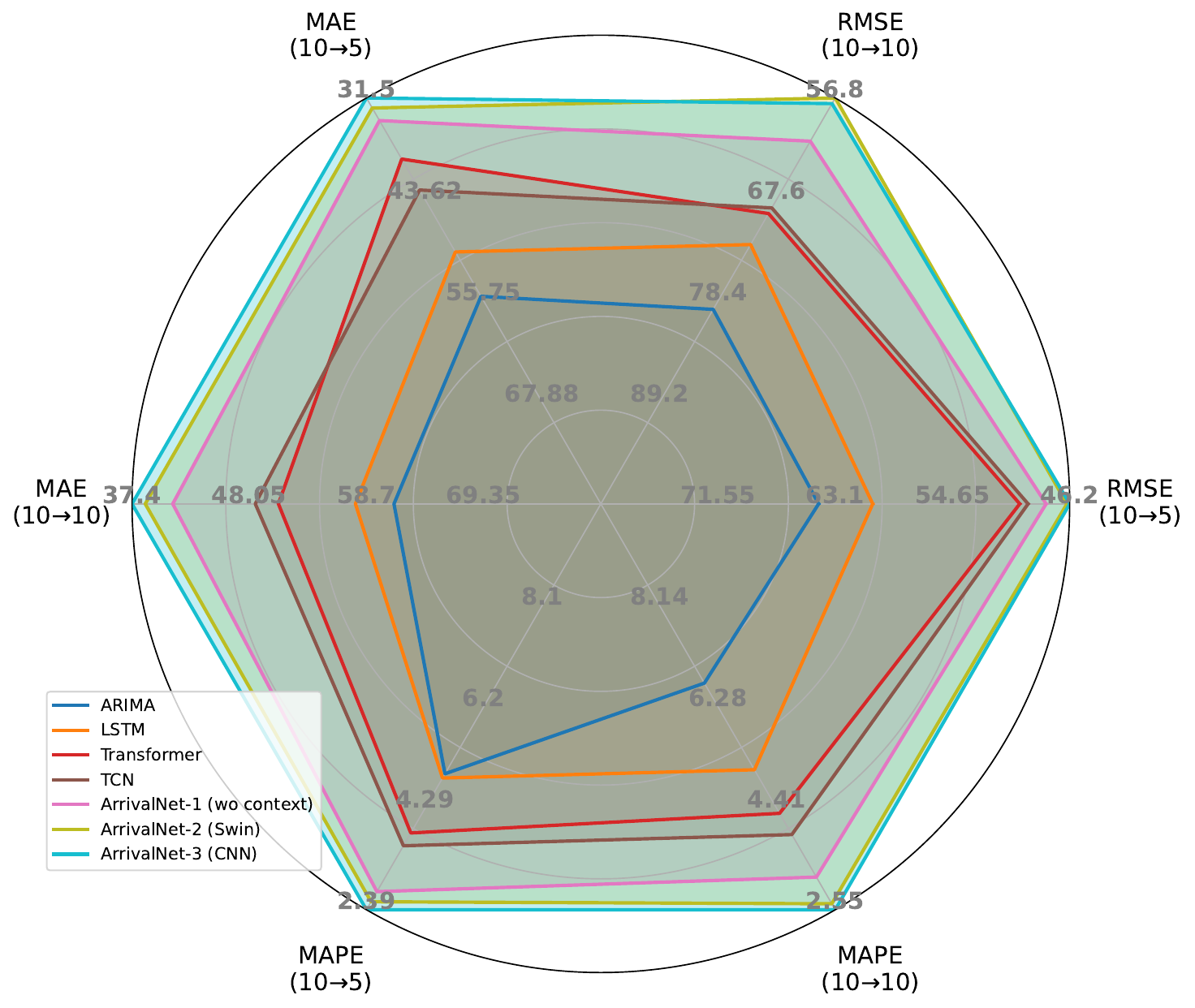}
    \caption{\textit{ArrivalNet}.}
    \label{fig_polar_tram}
  \end{subfigure}
  \begin{subfigure}[b]{0.48\textwidth}
    \includegraphics[width=1\textwidth]{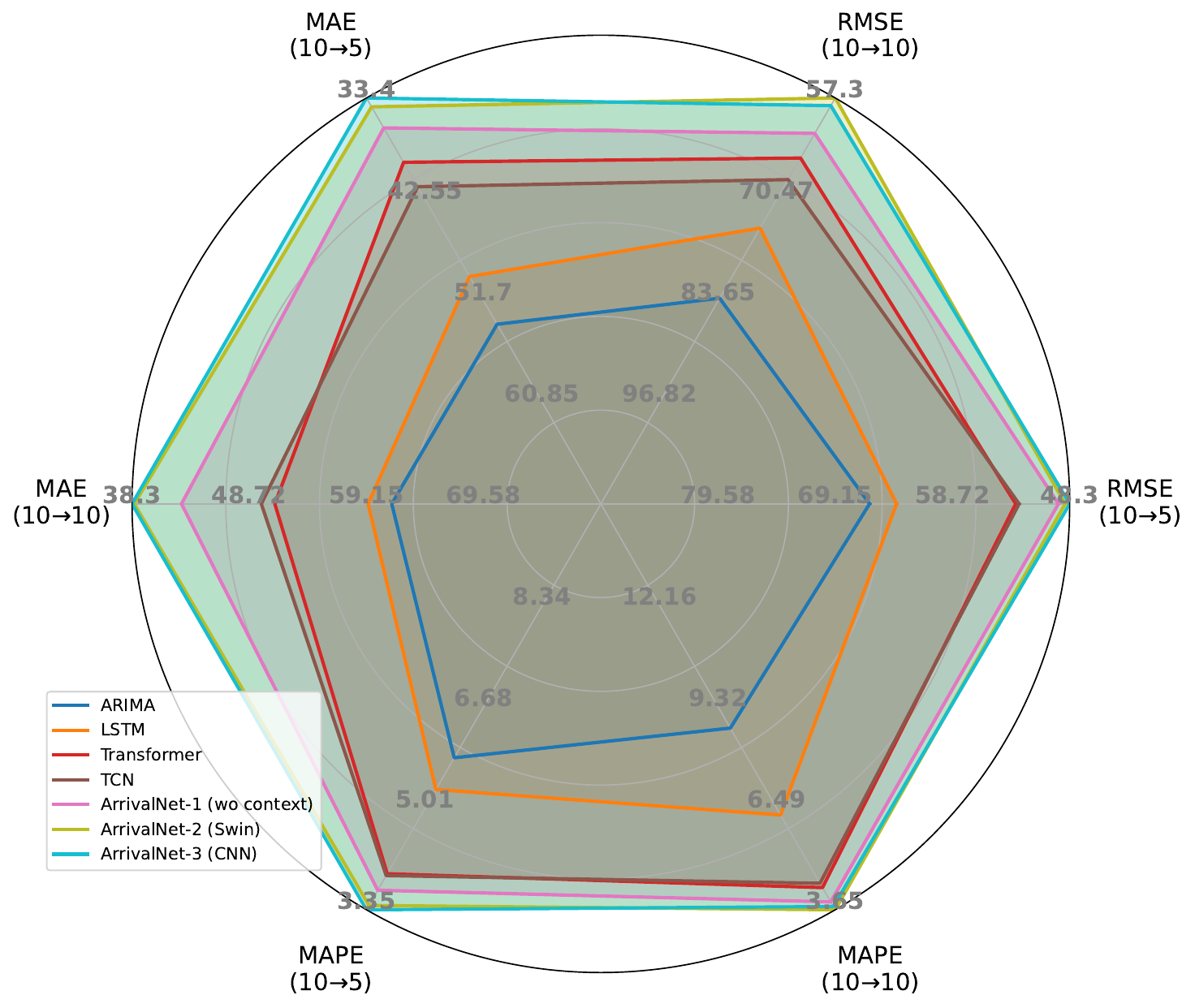}
    \caption{Transformer.}
    \label{fig_polar_bus}
  \end{subfigure}
  \caption{Radar charts for tram and bus with different metrics and experimental settings.}
    \label{fig_polar}
\end{figure*}

\begin{figure*}[h!]
  \centering
  \begin{subfigure}[b]{1\textwidth}
    \includegraphics[width=1\textwidth]{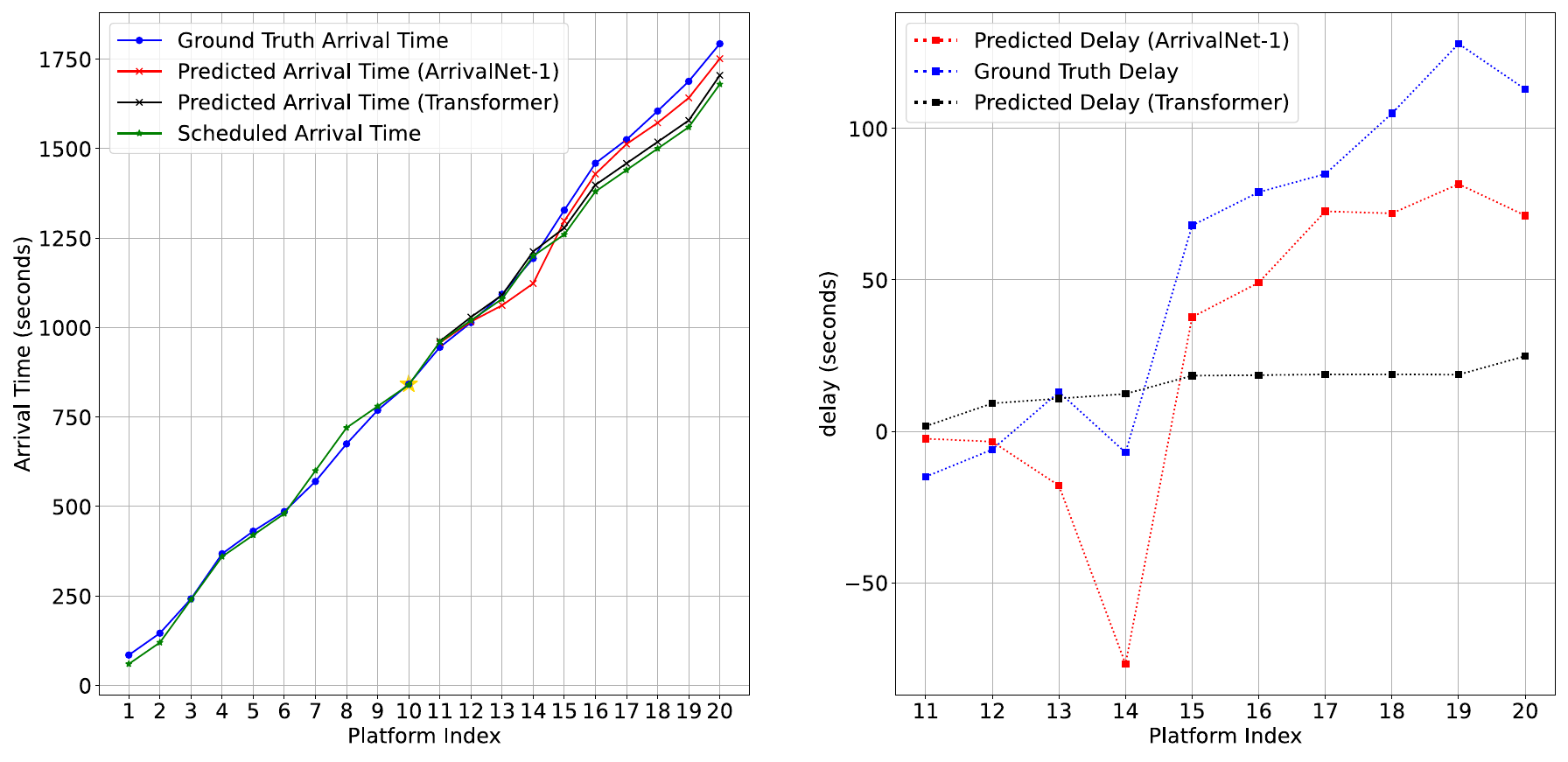}
    \caption{Tram case 1.}
    \label{tram_case_1}
  \end{subfigure}
 
  \begin{subfigure}[b]{1\textwidth}
    \includegraphics[width=1\textwidth]{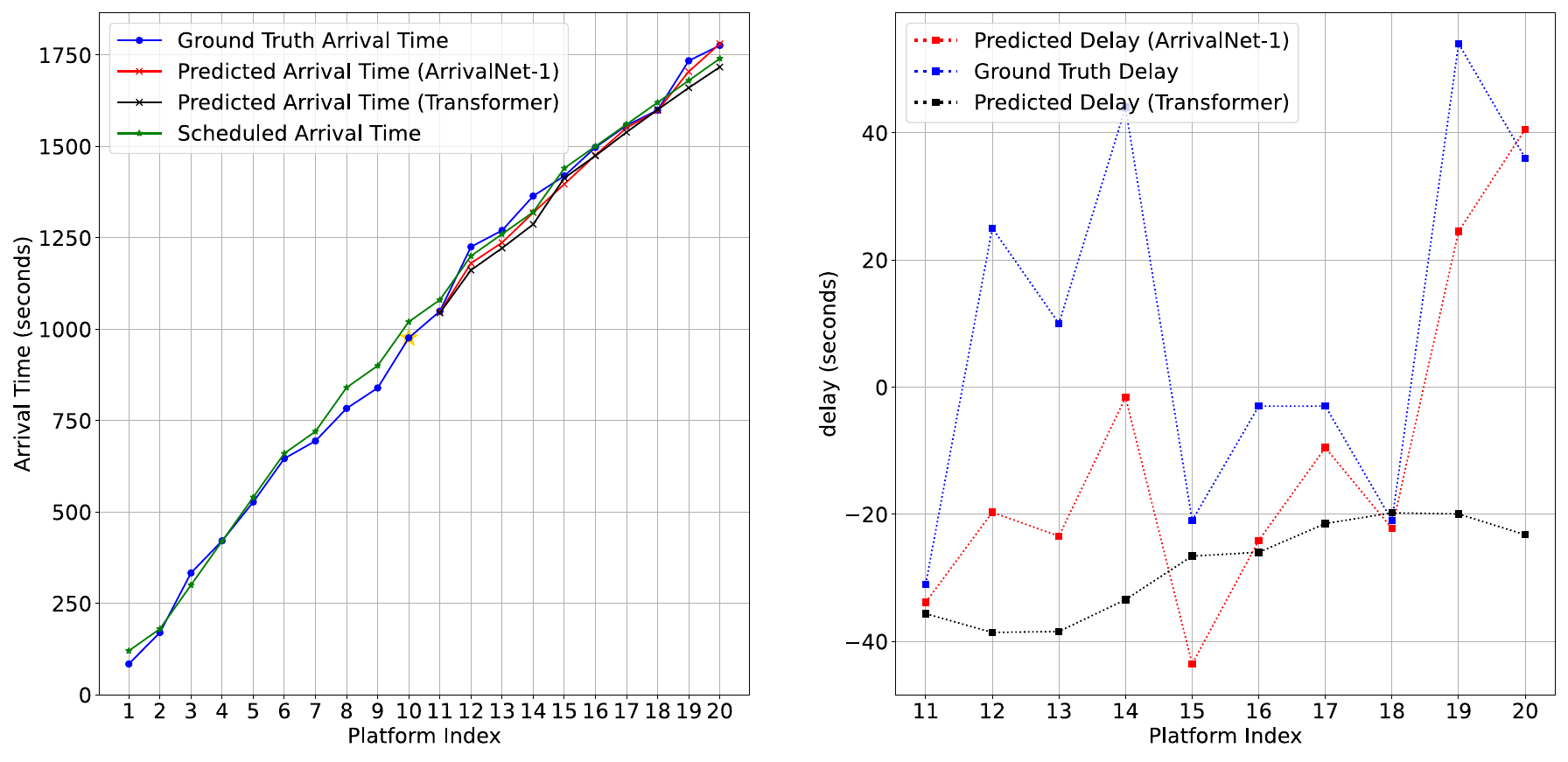}
    \caption{Tram case 2.}
    \label{tram_case_2}
  \end{subfigure}
  \caption{Two cases of tram ATP.}
  \label{tram_casestudy}
\end{figure*}

\begin{figure*}[h!]
  \centering
  \begin{subfigure}[b]{1\textwidth}
    \includegraphics[width=1\textwidth]{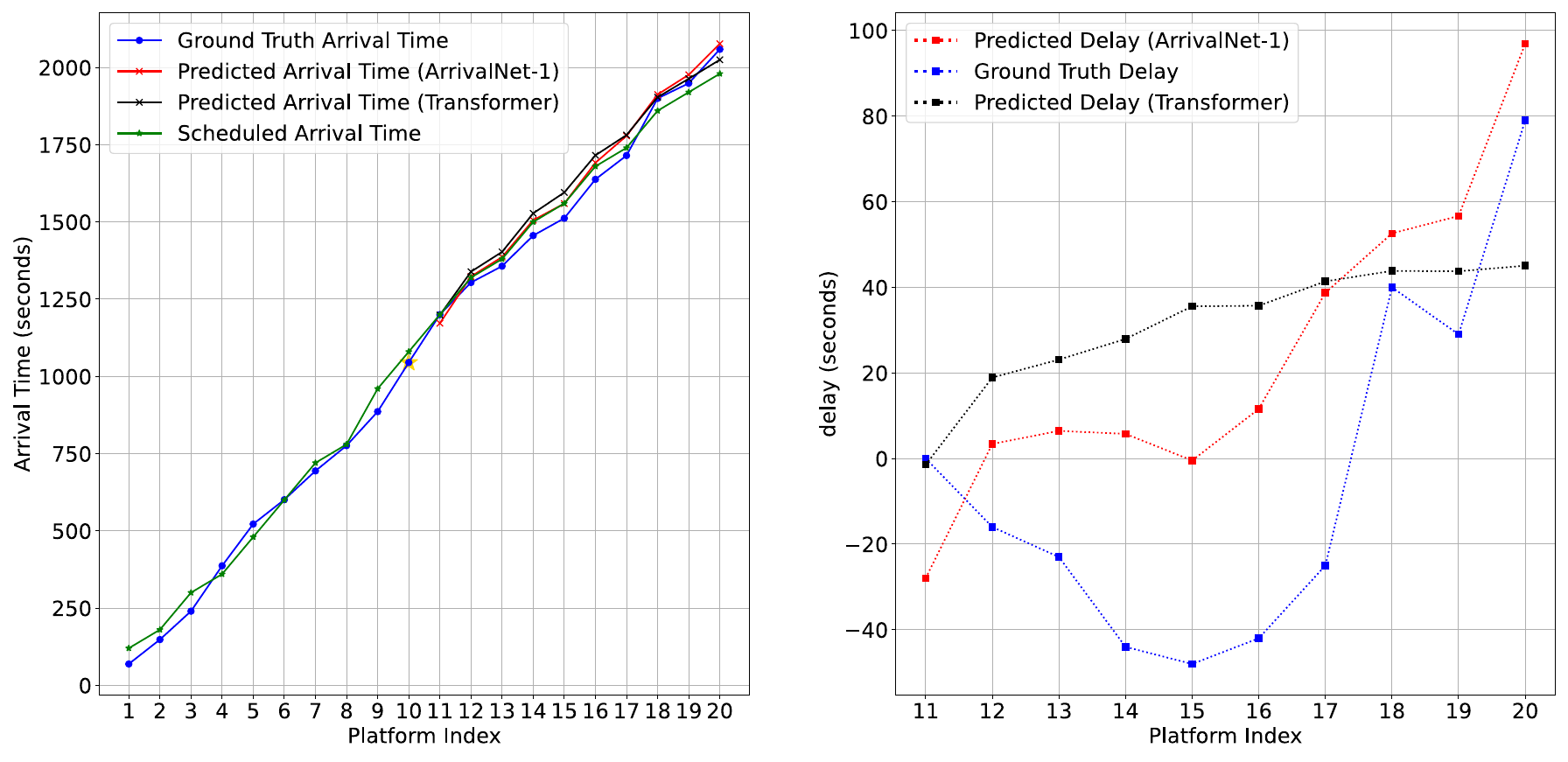}
    \caption{Bus case 1.}
    \label{bus_case_1}
  \end{subfigure}
 
  \begin{subfigure}[b]{1\textwidth}
    \includegraphics[width=1\textwidth]{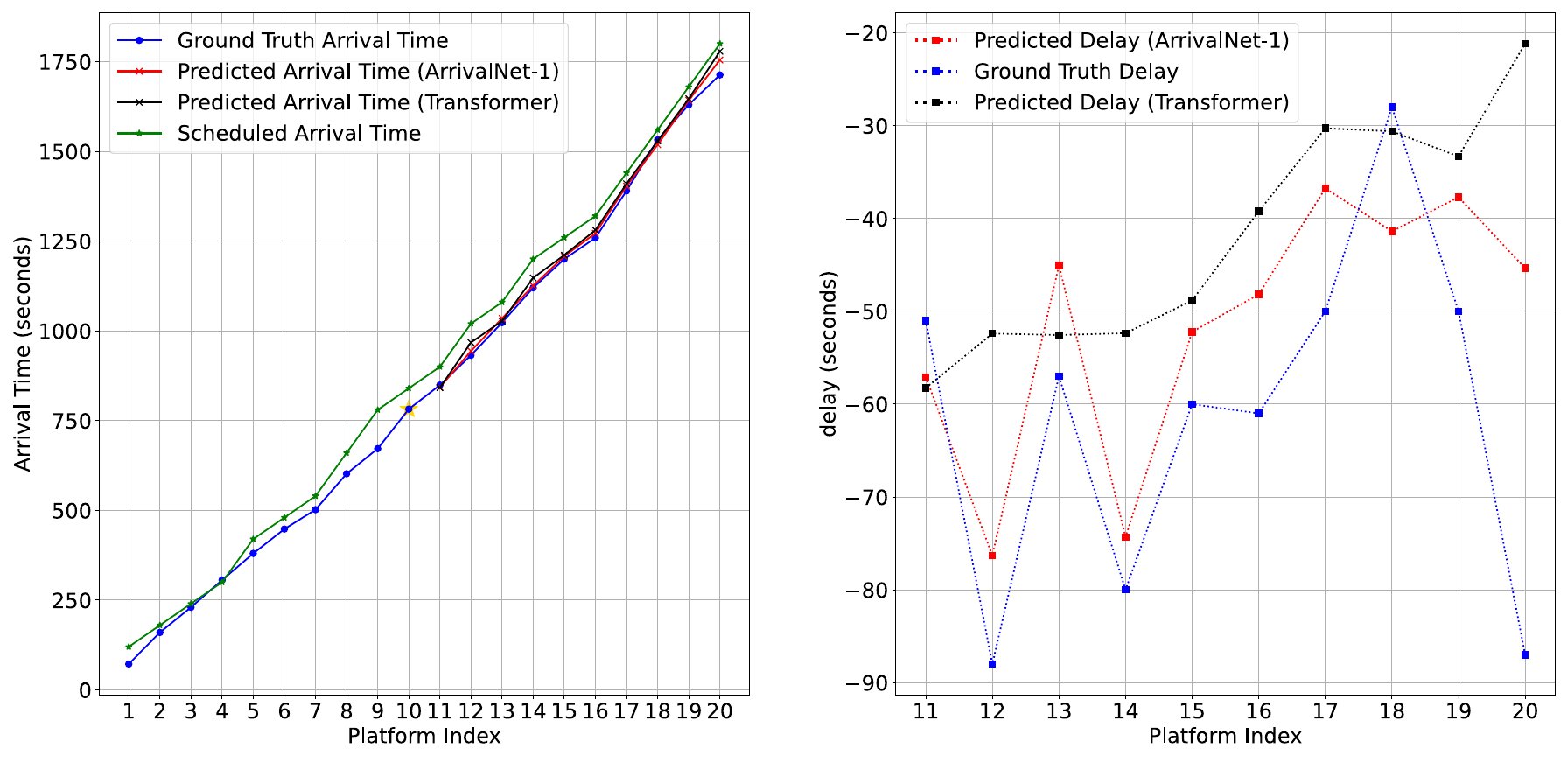}
    \caption{Bus case 2.}
    \label{bus_case_2}
  \end{subfigure}
  \caption{Two cases of bus ATP.}
    \label{bus_casestudy}
\end{figure*}

\subsection{Experimental settings, metrics and baselines}
In the training process, 90\% samples ($\sim$4.47M sequences) in the dataset are randomly selected as the training set, and the left samples ($\sim$0.5M sequences) are the testing set. The mean square error (MSE) is chosen as the loss function of \textit{ArrivalNet}. The details of parameter settings are presented in Table~\ref{table_parameters}. For \textit{ArrivalNet} and all baseline algorithms, the learning rate is set as 0.001, which is fit for the convergence of deep neural network. Following the hyperparameters in~\cite{wu2022timesnet}, the length of embedded feature, the number of 2D block and the number of kernels in CNN are set as 16, 2 and 6, respectively. Considering that most of valid sequence length is between 15 and 20, the input series length $N_{p}$ is set as 10 and the output lengths $N_{f}$ are 5 and 10. It can reflect the influence of sequence length to the model performance. The minimum length of the whole sequence is 15, which can be decomposed into 4 frequencies. Therefore, the 3 top frequencies are selected. The minimum width of the 2D tensor is 2, which is equal to the local window size. It can guarantee the effectivity of basic operation in the Swin Transformer. As for the threshold in related signal selection, it is the maximum distance between the location of signal infrastructure and the nearest waypoint on the link. If the distance to the  nearest waypoint  is less than the threshold, it is judged as the related signal. Experimentally, nearly all signal infrastructure can be captured and paired with the corresponding platform when the threshold is set to 20 m.

Three metrics are chosen for model performance evaluation, root mean square error (RMSE), mean absolute error (MAE), and mean absolute percentage error (MAPE). Compared to MAE, RMSE is more sensitive to large errors in prediction by squaring the errors before averaging. MAPE is useful in expressing errors as a proportion of actual values. At stop $i_k$, the formulations of three metrics are detailed as follows:
\begin{equation}
\text{RMSE} = \sqrt{\frac{1}{N_{f}}\sum_{t=i_{k}+1}^{i_{k}+N_{f}}  (\mathbf{T}^{\text{arrival}}_{i_{k}+t} - \hat{\mathbf{T}}^{\text{arrival}}_{i_{k}+t})^2}
\end{equation}
\begin{equation}
\text{MAE} = \frac{1}{N_{f}}\sum_{t=i_{k}+1}^{i_{k}+N_{f}} \left| \mathbf{T}^{\text{arrival}}_{i_{k}+t} - \hat{\mathbf{T}}^{\text{arrival}}_{i_{k}+t} \right|
\end{equation}
\begin{equation}
\text{MAPE} = \frac{100}{ N_{f}}\sum_{t=i_{k}+1}^{i_{k}+N_{f}} \left| \frac{\mathbf{T}^{\text{arrival}}_{i_{k}+t} - \hat{\mathbf{T}}^{\text{arrival}}_{i_{k}+t}}{\mathbf{T}^{\text{arrival}}_{i_{k}+t}} \right|
\end{equation}
where $N_{f}$ is number of future steps. $i_{k}$ is the stop index of $k^{th}$ sequence. $\hat{\mathbf{T}}^{\text{arrival}}_{i_{k}+t}$ and $\mathbf{T}^{\text{arrival}}_{i_{k}+t}$ are estimation and groundtruth values of $k^{th}$ testing sequences at $(i_{k}+t)^{th}$ stop, respectively. 

To validate the performance of \textit{ArrivalNet}, a few baseline methods are implemented for comparison, including the traditional time series smoothing method, the RNN-based method, the attention-based method, and the CNN-based method. Meanwhile, three variants of \textit{ArrivalNet} are designed for comparative study. An overview of each method is as follows:
\begin{itemize}
    \item \textbf{Traditional time series smoothing method}. Auto-regressive integrated moving average (ARIMA) is selected as the traditional smoothing method, which is a widely used statistical approach for time series forecasting~\cite{box1970distribution}. It  is a combination of the differenced autoregressive (AR) model with the moving average (MA) model.  The AR part of ARIMA shows that the time series is regressed on its own past data. The integrated part is used to make the time series stationary by differencing the observations a certain number of times. The MA part indicates that the forecast error is a linear combination of past respective errors. 
    \item  \textbf{RNN-based method}. LSTM~\cite{hochreiter1997long}, as a typical recurrent network, is chosen for comparison. LSTM enhances the representation capability for one-dimensional temporal processes by modeling both long and short-term memories. It has been widely applied in the prediction of public transport arrival times~\cite{Ratneswaran,rong2022bus}.
    \item  \textbf{Attention-based method}. Transformer is a well-developed, attention-based method and has been widely applied in natural language processing and trajectory prediction~\cite{vaswani2017attention}. Due to the attention mechanism, correlations between non-adjacent time points in long temporal sequences can be captured.
    \item  \textbf{CNN-based method}. Temporal convolutional network (TCN) is a typical time series method developed based on CNN~\cite{bai2018empirical}. It captures temporal correlations through convolutional kernels along the time dimension, which has also been widely used in sequential human action recognition and trajectory prediction~\cite{mohamed2020social}.
    \item \textbf{\textit{ArrivalNet-1}} is the \textit{ArrivalNet} without considering the contextual features, which is equipped with CNN vision backbone. 
    \item \textbf{\textit{ArrivalNet-2}} is the \textit{ArrivalNet} with Swin Transformer vision backbone. 
    \item \textbf{\textit{ArrivalNet-3}} is the \textit{ArrivalNet} with CNN vision backbone.
\end{itemize}
\subsection{Results}
\subsubsection{Quantitative results}
The results regarding the ATP of tram and bus are presented in Tables~\ref{table_tram} and Table~\ref{table_bus}. In each table, three metrics (RMSE, MSE, and MAPE) are shown for two different output lengths (10$\rightarrow$5 and 10$\rightarrow$10). In Table~\ref{table_tram}, three variants of proposed methods (\textit{ArrivalNet-1}, \textit{ArrivalNet-2} and \textit{ArrivalNet-3}) outperform other baselines across all three metrics. Among three \textit{ArrivalNet}, the absence of contextual features leads to a decrease in performance, while \textit{ArrivalNet-2} or \textit{ArrivalNet-3} achieve the best results in all six comparsions. In the tram ATP, compared to the four baseline methods (average of two lengths), the best-performing \textit{ArrivalNet} model reduces RMSE by 28.26\% (51.6 vs 71.8), 22.29\% (51.6 vs 66.4), 13.20\% (51.6 vs 59.45), 12.32\% (51.6 vs 58.85), MAE by 40.81\% (34.45 vs. 58.2), 35.97\% (34.45 vs. 53.8), 23.02\% (34.45 vs. 44.75), 24.37\% (34.45 vs. 45.55), and MAPE by 57.59\% (2.47 vs. 5.825), 50.50\% (2.47 vs. 4.99), 39.39\% (2.47 vs. 4.075), 34.31\% (2.47 vs. 3.76).  In four baseline methods,  for each metric, the performance of three deep learning methods (LSTM, Transformer, TCN) surpasses that of the autoregressive method (ARIMA), indicating that deep learning-based approaches have advantages in sequential prediction. 

Similar to Table~\ref{table_tram}, Table~\ref{table_bus} presents the corresponding results for bus ATP. In the bus ATP, for the average of each metric in two lengths, the best-performing \textit{ArrivalNet} model reduces RMSE by 29.05\% (53 vs. 74.7), 23.13\% (53 vs. 68.95), 10.32\% (53 vs. 59.1), 12.18\% (53 vs. 60/35), MAE by 37.76\% (35.85 vs. 57.6), 34.10\% (35.85 vs. 54.4), 20.51\% (35.85 vs. 45.1), 21.38\% (35.85 vs. 45.6), MAPE by 51.48\% (3.535 vs. 7.285), 39.16\% (3.535 vs. 5.81), 13.89\% (3.535 vs. 4.105), 14.82\% (3.535 vs. 4.15). The findings from Table~\ref{table_bus} are similar to those of Table~\ref{table_tram}. Overall, Table~\ref{table_tram} and Table~\ref{table_bus} indicate that \textit{ArrivalNet} achieves the best performance in both tram and bus ATP. To more clearly illustrate the performance comparison among all algorithms, Fig.~\ref{fig_polar} presents two radar charts for tram and bus across two length settings and three metrics. The radar charts also demonstrate that the proposed \textit{ArrivalNet} consistently achieves the best performance across different metrics and experimental settings.

\begin{figure}[t]
  \centering
  \includegraphics[width=0.48\textwidth]{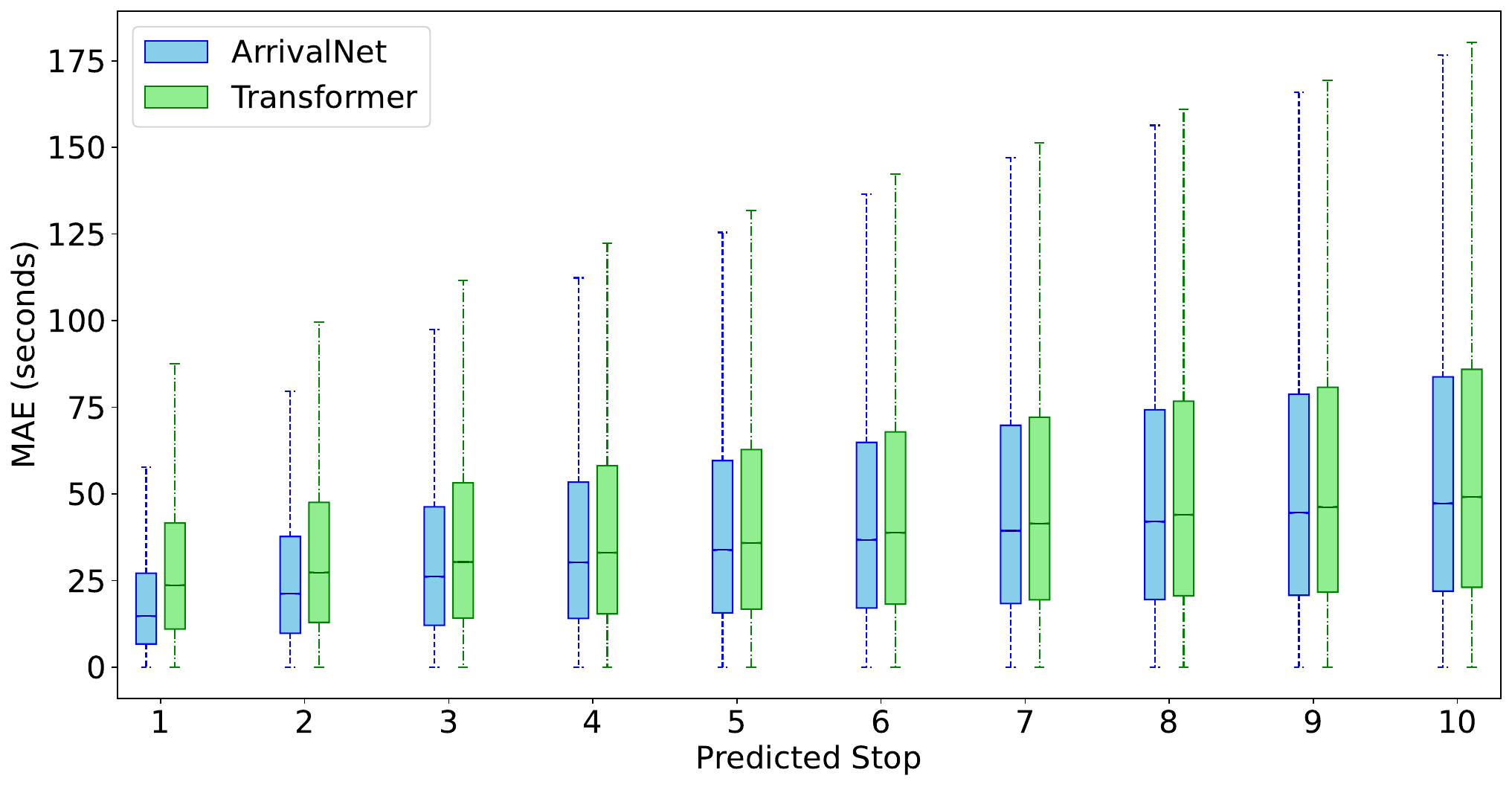}
  \caption{The boxplot of MAE for each predicted stop. (vehicle type: tram, prediction length: 10$\rightarrow$10)}
  \label{fig_tram_box_plot}
\end{figure}

\begin{figure}[htbp]
  \centering
  \begin{subfigure}[b]{0.48\textwidth}
    \includegraphics[width=1\textwidth]{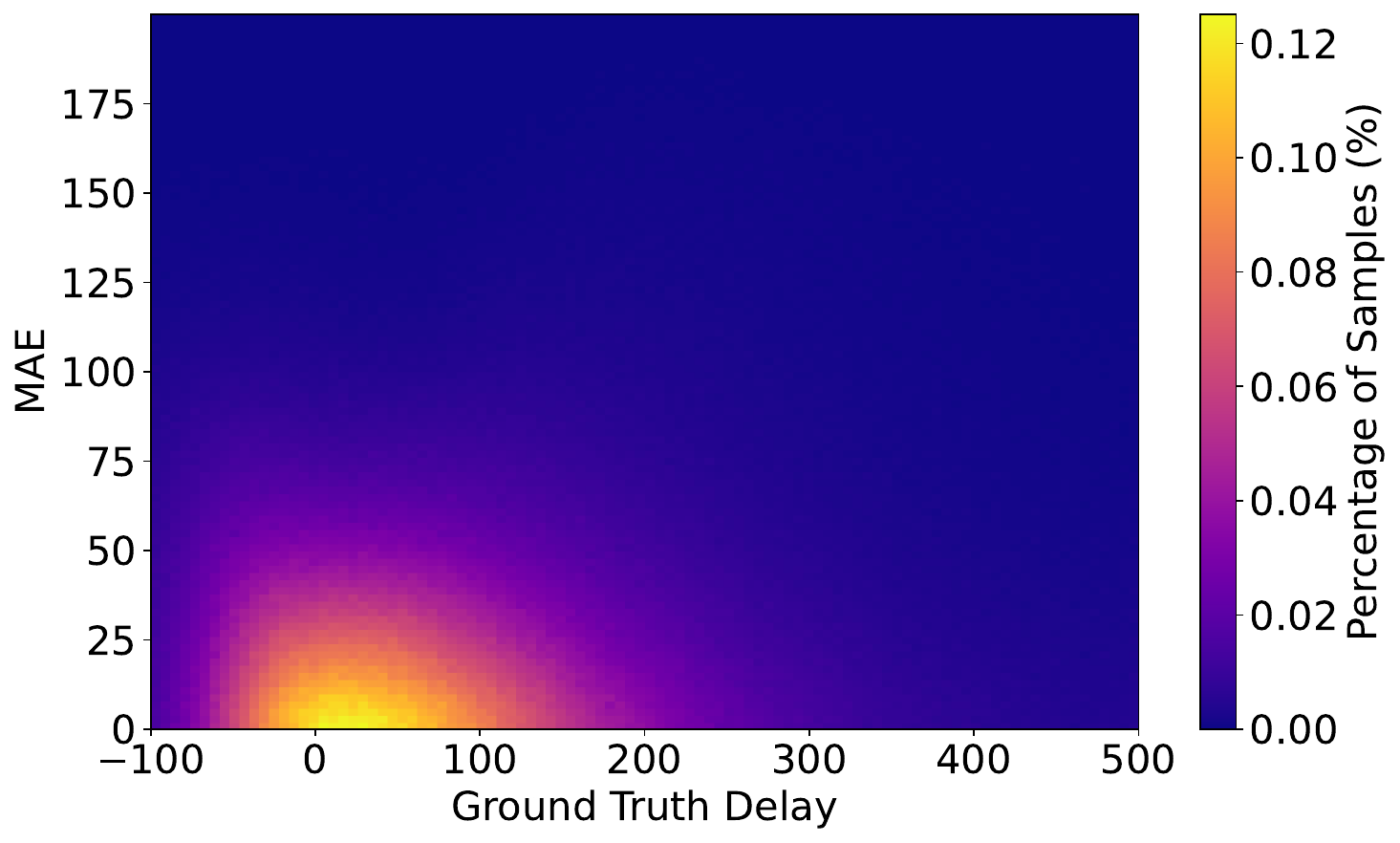}
    \caption{\textit{ArrivalNet}.}
    \label{fig_relation_1}
  \end{subfigure}
  
  \begin{subfigure}[b]{0.48\textwidth}
    \includegraphics[width=1\textwidth]{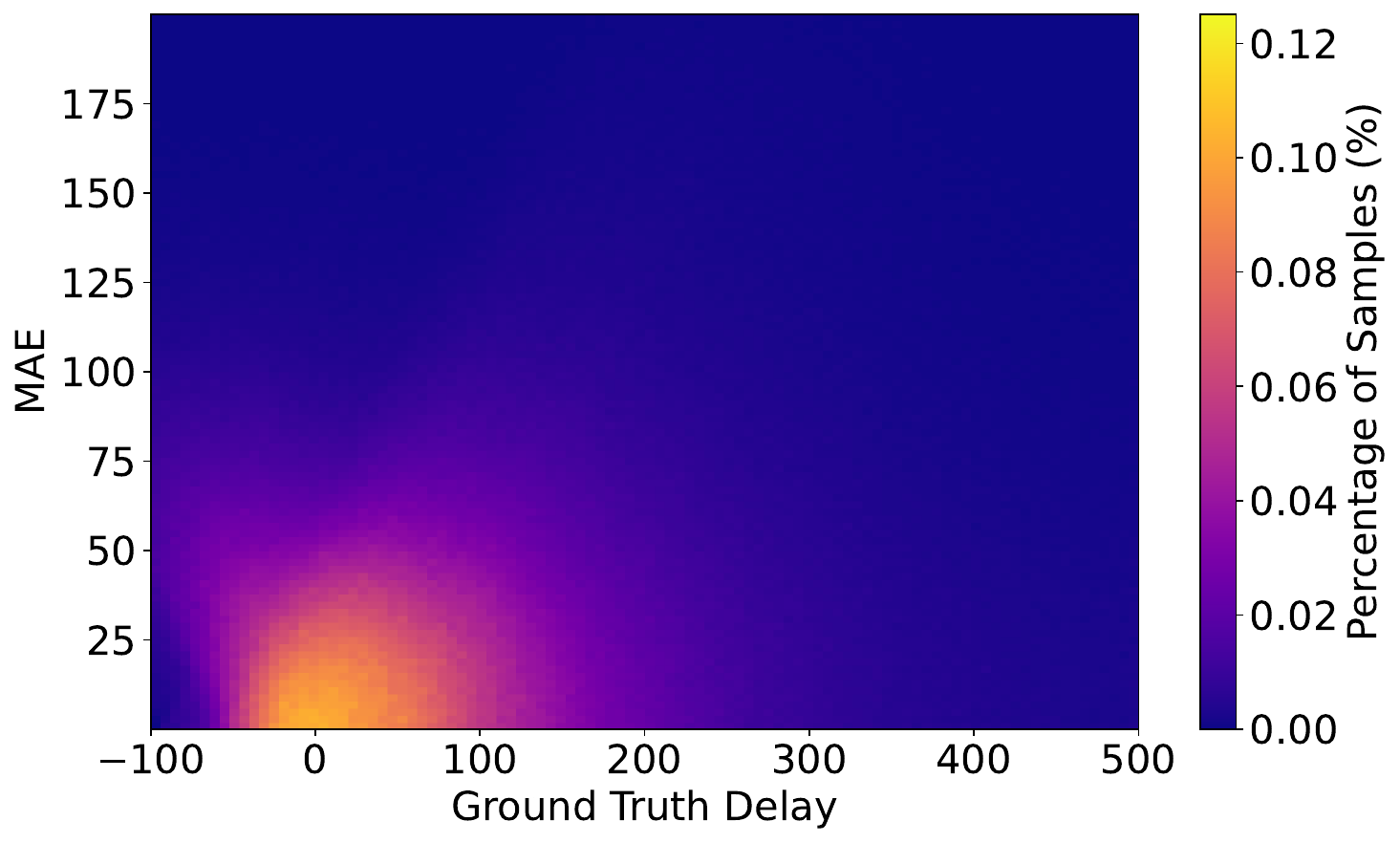}
    \caption{Transformer.}
    \label{fig_relation_2}
  \end{subfigure}
  \caption{The relationship of between MAE and ground-truth delays (vehicle type: tram, prediction length: 10$\rightarrow$10).}
    \label{fig_relation}
\end{figure}

\subsubsection{Case study}
Fig.~\ref{tram_casestudy} and Fig.~\ref{bus_casestudy} present several cases of tram and bus ATP. Each case is a sequential arrival information of tram or bus with 20 stops. The labels of the x axis only represents the platform index of the corresponding sequence, it indicates that the same index in different cases doesn't mean the same physical public transport stop. In all cases, the length of both input and output sequences are set to 10, corresponding to platforms 1-10 and 11-20, respectively. In all cases, the vehicle departs from the $10^{th}$ platform, which is highlighted with a golden star. In each figure, the left one is the comparison of ground truth, predicted and scheduled arrival time, while the right one is the comparsion of  ground truth, predicted delay. Considering that Transformer is the method with best performance in all one-dimensional algorithms overall, it is selected for the comparative study. 

For all cases of trams and buses, the scheduled arrival time do not match the ground truth arrival time, indicating that discrepancies between actual arrival times and timetables are common. It emphasizes the significance of public transport ATP. Comparing the ground truth delay, predicted delay (\textit{ArrivalNet}) and predicted delay (Transformer) across all cases, \textit{ArrivalNet} is good at capturing the small variations in the sequence, such as rising, falling, fluctuation. For example, the platforms 13-15 in tram case 1, the platforms 14-18 in tram case 2, the platforms 18-20 in bus case 1 and the platforms 11-15 in bus case 2. On the contrary, Transformer cannot respond to these changes promptly, leading to only minor increases or decreases in the prediction of delay. It tends to generate an average of multi-step prediction. 

\begin{figure}[ht!]
  \centering
  \includegraphics[width=0.48\textwidth]{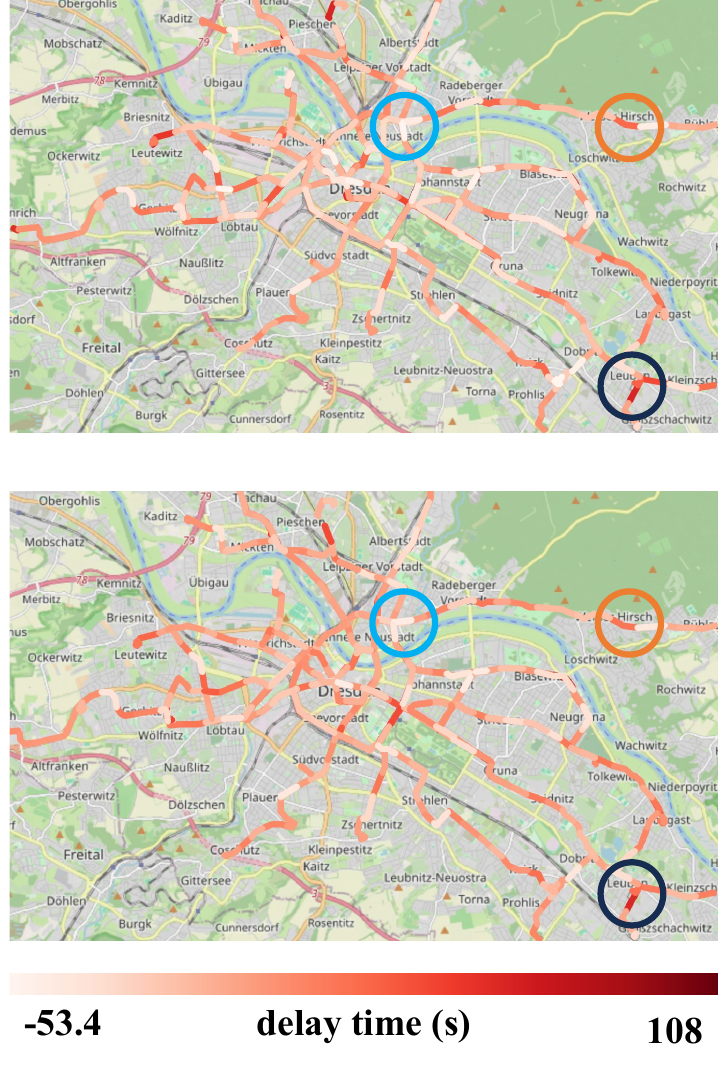}
  \caption{The city-wide tram link delay distributions (top: ground-truth link delay distribution, bottom: predicted link delay distribution from \textit{ArrivalNet-3}).}
  \label{fig_tram_delay_dist}
\end{figure}

\begin{figure}[ht!]
  \centering
  \includegraphics[width=0.48\textwidth]{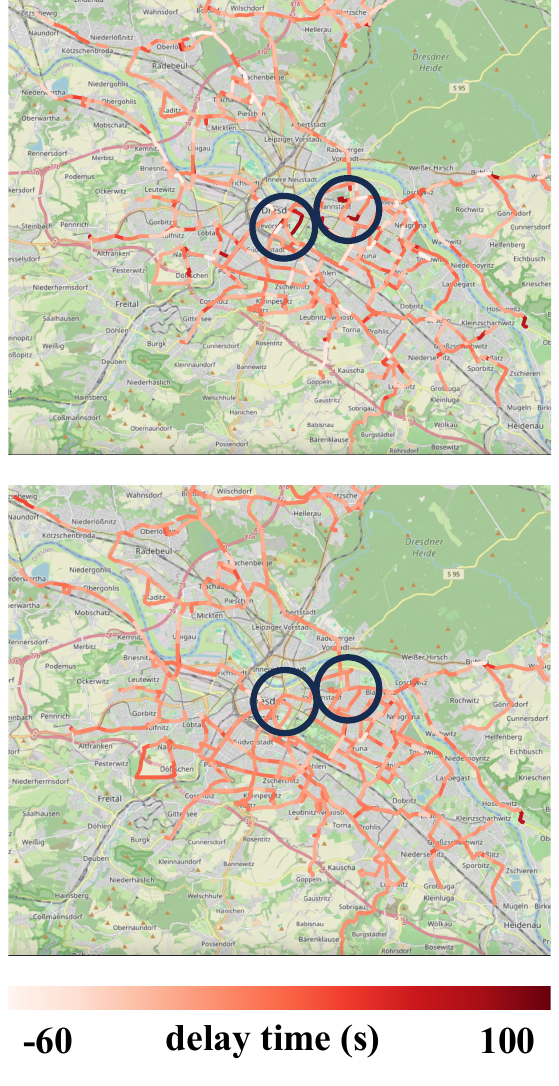}
  \caption{The city-wide bus link delay distributions (top: ground-truth link delay distribution, bottom: predicted link delay distribution from \textit{ArrivalNet-3}).}
  \label{fig_bus_delay_dist}
\end{figure}

\subsubsection{Statistical analysis}
Fig.~\ref{fig_tram_box_plot} shows the MAE boxplot of all test samples ($\sim$0.5M sequences) for different predicted platforms by \textit{ArrivalNet-2} and Transformer. The input and output lengths set to 10. For both methods, MAE increases as the prediction length extends, which aligns with the general trend observed in time series prediction. \textit{ArrivalNet} achieves a smaller MAE across all predicted platforms, including the mean and variance. Moreover, this difference is more pronounced for nearer platforms (predicted platforms 1, 2, 3).

To better understand the advantage of \textit{ArrivalNet} in ATP, Fig.~\ref{fig_relation_1} illustrates the joint  distribution between MAE and different delays of all test samples. In this figure, the horizontal axis represents the actual delay at each platform, while the vertical axis corresponds to MAE. The color of each grid represents the percentage of samples in the test set. Comparing the results of \textit{ArrivalNet} and Transformer, MAE of \textit{ArrivalNet} is predominantly concentrated in intervals with smaller values. As the value of delay increases, unlike \textit{ArrivalNet}, MAE of Transformer is not concentrated in a smaller interval but tend to disperse over a broader range. This comparison suggests that the proposed \textit{ArrivalNet} can maintain predicted errors within a smaller range.

\subsubsection{The city-wide analysis}
In this work, city-wide operational data is utilized. The delay in extracted sequences is the cumulative delay, which is influenced by the delay propagation along the route. To analyze the distribution of actual and predicted delays in city-wide public transport, the cumulative delay at different platforms along the same route is converted into the link delay, which represents the delay caused by tram/bus between two adjacent platforms. The transformation from cumulative to link delays is to calculate the difference of cumulative delay between adjacent platforms. As shown in Fig.~\ref{fig_tram_delay_dist}, the comparison of actual and predicted delays shows that the predictions of \textit{ArrivalNet} for trams are generally consistent with ground truth. For example, in the three pairs of circle, the proposed \textit{ArrivalNet} successfully capture the high (black), low (orange) and negative (blue) delays. It indicates that for tram/bus ATP (delay prediction), \textit{ArrivalNet} can accurately capture the real city-wide delay distribution and pattern. 
\begin{table}[htbp]
\captionsetup{font={small}}
\caption{The influence of negative delay (vehicle type: tram, prediction length: 10)}
\centering
\begin{tabular}{cccc}
\toprule
\textbf{metric} & w/ negative delay & w/o negative delay & error increase \\
\hline
RMSE & 57.4 & 59.78 & 4.0\% \\
MAE & 37.4 & 39.1 & 4.55\% \\
MAPE & 2.55 & 2.84 & 11.37\% \\
\bottomrule
\end{tabular}
\label{table_negative_delay}
\begin{tablenotes}
\item w/ negative delay: Same with previous experiments in Table.~\ref{table_tram} and Table.~\ref{table_bus}.
\item w/o negative delay: All negative delay in samples are set as 0.
\end{tablenotes}
\end{table}

\begin{figure}[ht!]
  \centering
  \begin{subfigure}[b]{0.48\textwidth}
    \includegraphics[width=1\textwidth]{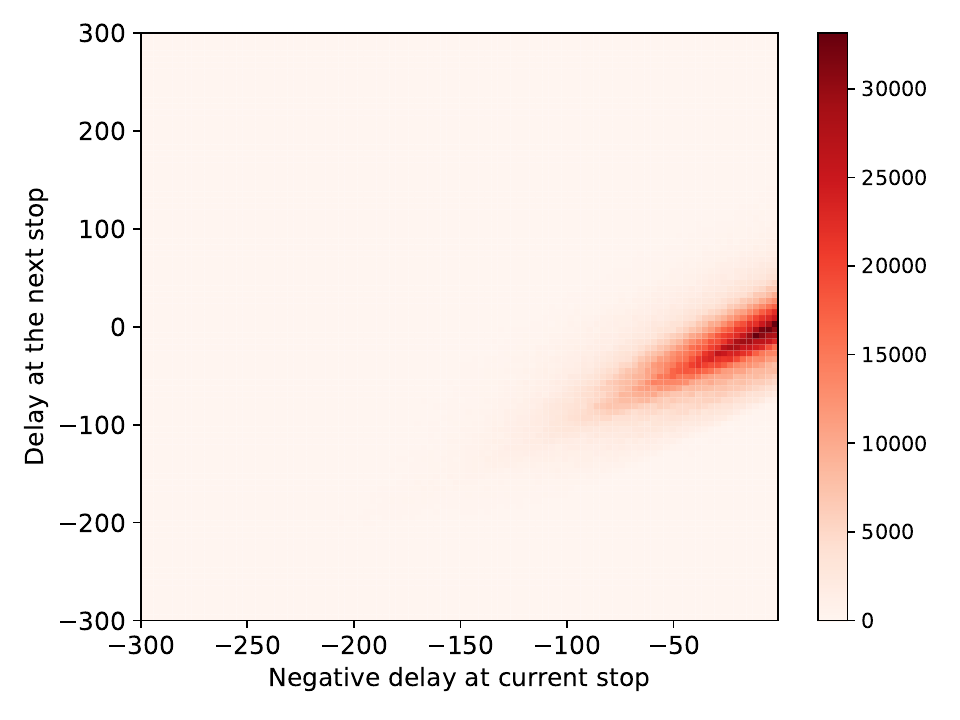}
    \caption{The 2D heatmap.}
    \label{fig_negative_delay_heatmap}
  \end{subfigure}
  
  \begin{subfigure}[b]{0.48\textwidth}
    \includegraphics[width=1\textwidth]{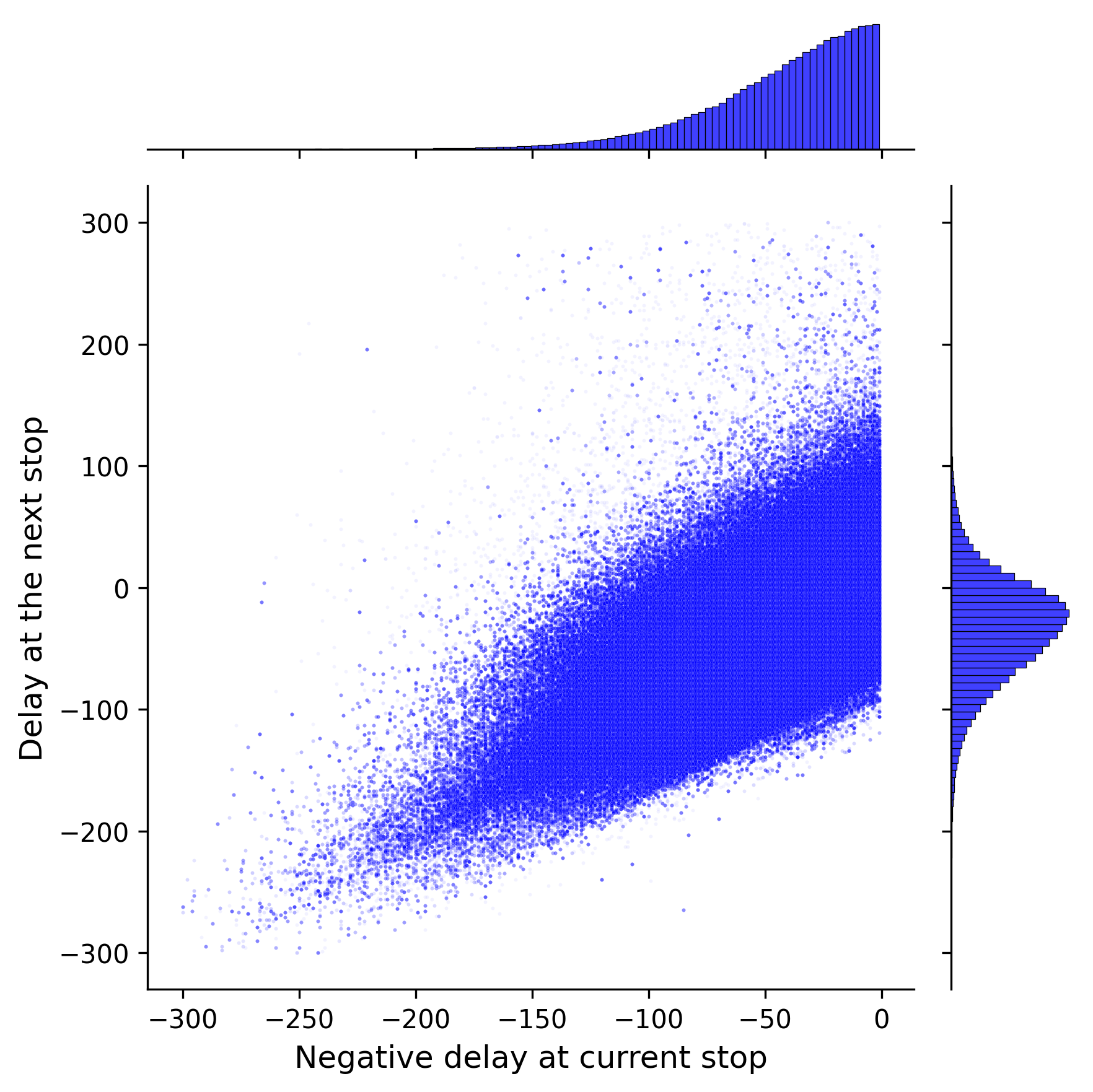}
    \caption{The 2D scatter distribution.}
    \label{fig_negative_delay_joint_distribution}
  \end{subfigure}
  \caption{The relationship of negative delay at the current stop and the delay at the next stop.}
    \label{fig_negative_delay}
\end{figure}

Similarly, Fig.~\ref{fig_bus_delay_dist} presents the city-wide distribution of the ground truth and the average predicted link delays. Comparing the ground truth link delay between the tram and the bus at the top of Fig.~\ref{fig_tram_delay_dist} and Fig.~\ref{fig_bus_delay_dist}, the overall delay of the bus is higher than that of the tram. In the comparison within Fig.~\ref{fig_bus_delay_dist}, it indicates that  \textit{ArrivalNet} can well predict the city-wide link delay distribution with low positive values. However, it also reveals that \textit{ArrivalNet} cannot perfectly capture the actual delay modes in downtown areas, which is also highlighted by the black circle. This could be due to the presence of more traffic congestion in downtown areas. Another possible reason is that most of Dresden's tram links use dedicated rail lines that do not overlap with roads used by buses and private cars. Therefore, the arrival time prediction of trams is less affected by the uncertainty of artery traffic conditions. This makes it easier to predict than buses.

\subsubsection{The influence of negative delay}
In this study, the Dresden city-wide data has instances with negative delay, indicating that the vehicle arrives at the stop earlier than the scheduled time. The causes of negative delay are varied, such as the absence of passengers boarding or alighting at specific stops, or bunching issues between consecutive vehicles. Typically, public transport drivers mitigate negative delays by increasing the dwell time at stops to prevent the propagation of the delay to subsequent stops. 

Fig.~\ref{fig_negative_delay} illustrates the relationship between delays at stops exhibiting negative delay and the delays at the subsequent stops. Specifically, Fig.~\ref{fig_negative_delay_heatmap} presents a 2D distribution of all samples in the dataset with negative delay, while Fig.~\ref{fig_negative_delay} shows a joint plot combining scatter and 1-dimensional distributions. From these figures, it is evident that overall, vehicles tend to reduce or eliminate negative delay at the next stop. Approximately 30.31\% of samples transition to a positive delay at the subsequent stop. This indicates that public transport vehicles actively work to reduce negative delays to maintain alignment with the schedule. Ideally, negative delays would not be propagated to the next stop, thereby preventing any impact on arrival time predictions.

To further analyze the impact of negative delay on the performance of ATP, an additional experiment was conducted. For the tram ATP, all negative delays were set to zero during model training. The results of this experiment are presented in Table~\ref{table_negative_delay}, where the length of the prediction step is 10. It shows that including negative delay helps the model learn a more accurate delay propagation process. Conversely, when negative delay is excluded, the model's performance slightly declines, though it does not result in a complete failure of the model. Future research may involve incorporating external knowledge to more precisely characterize the propagation of negative delay, thereby enhancing the training process of the deep learning model.

\begin{figure*}[ht!]
  \centering
  \includegraphics[width=1\textwidth]{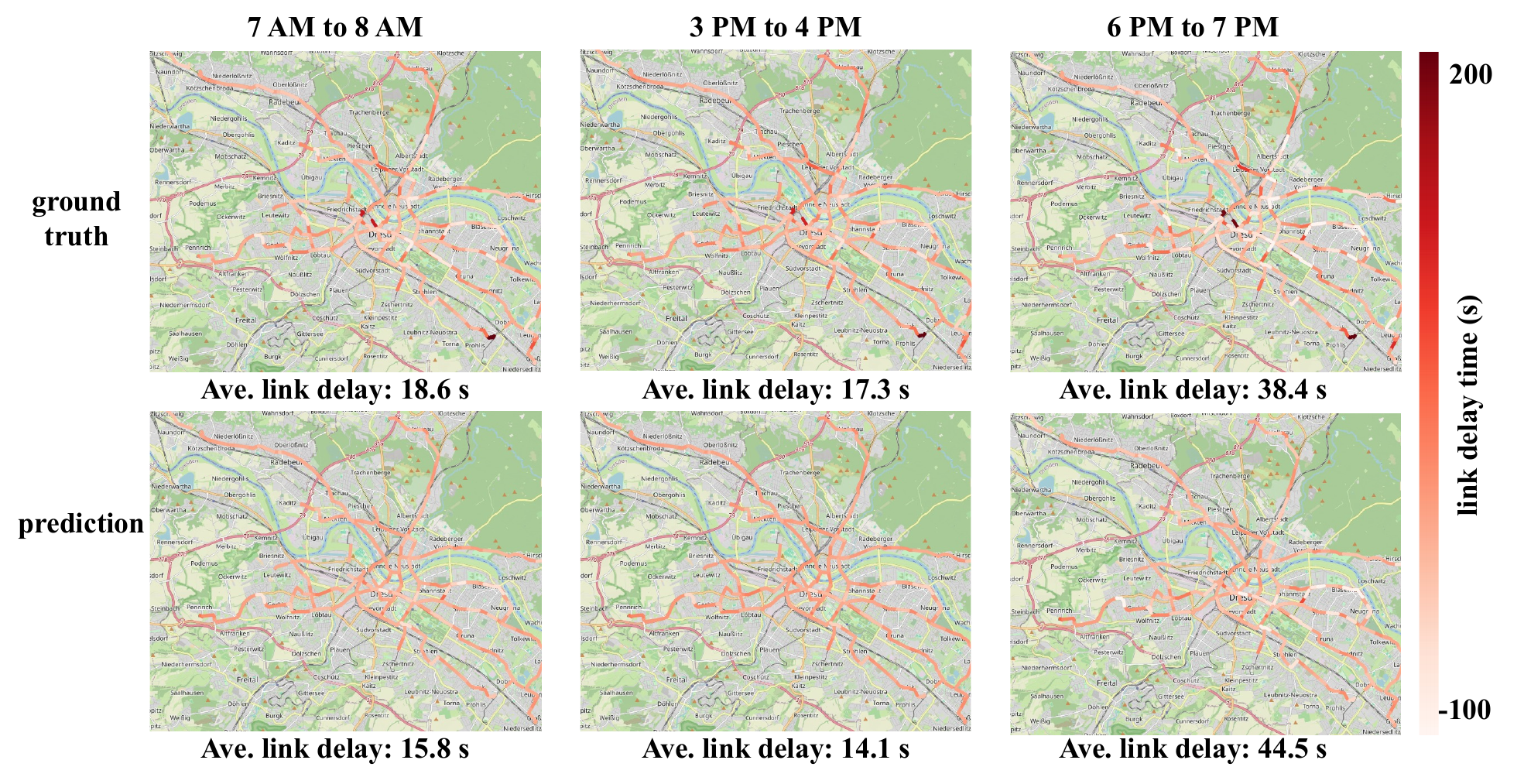}
  \caption{The city-wide hour-level tram link delay distributions (top row: ground-truth link delay distribution, bottom row: predicted link delay distribution from \textit{ArrivalNet-3}).}
  \label{fig_tram_hour_distribution}
\end{figure*}

\begin{figure*}[ht!]
  \centering
  \includegraphics[width=1\textwidth]{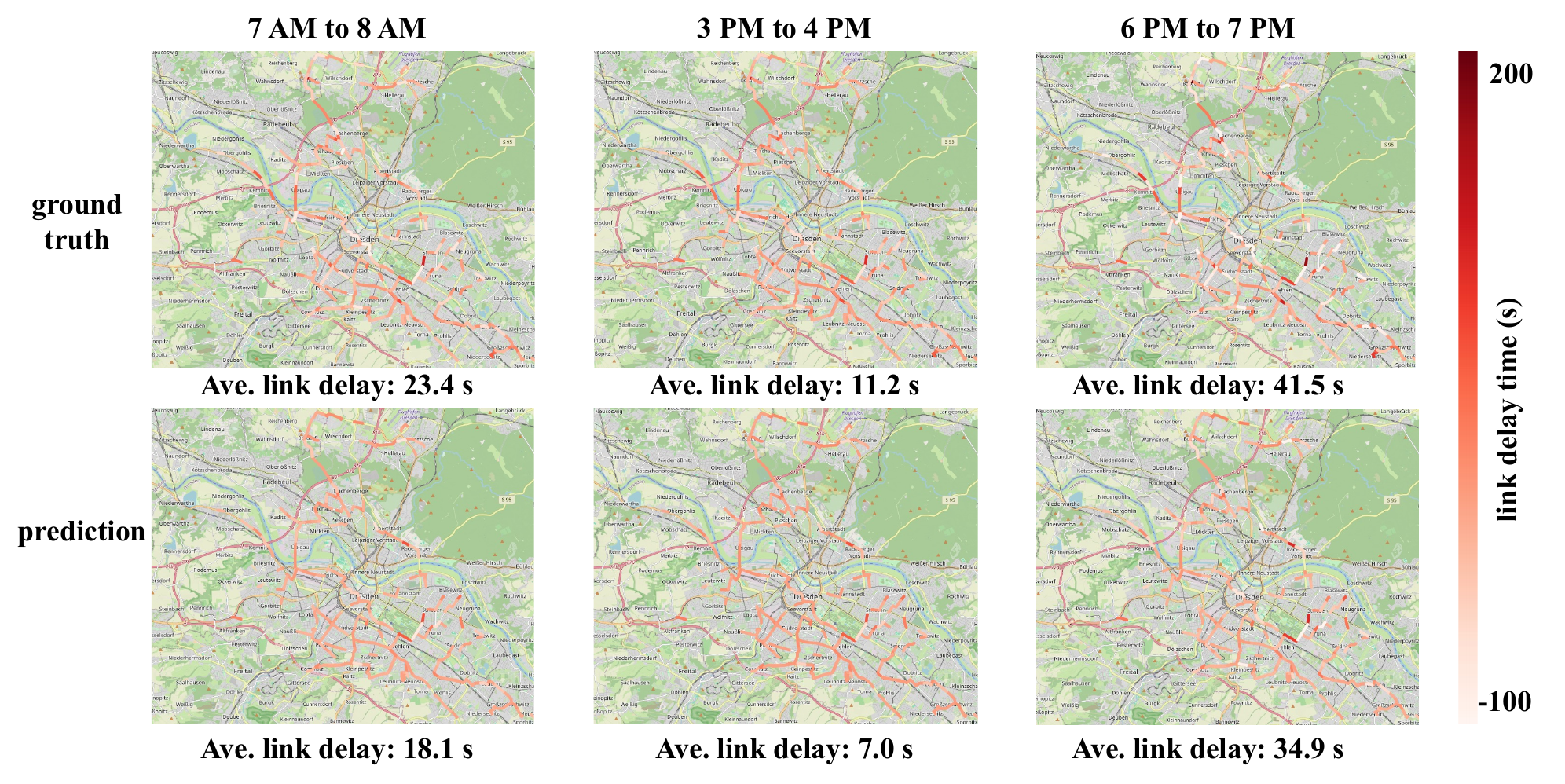}
\caption{The city-wide hour-level bus link delay distributions (top row: ground-truth link delay distribution, bottom row: predicted link delay distribution from \textit{ArrivalNet-3}).}
  \label{fig_bus_hour_distribution}
\end{figure*}

In addition to analyzing the city-wide link delay distribution of all collected data, this work also presents the variation in city-wide link delay at each hour of the day to investigate the performance of \textit{ArrivalNet}'s estimation at the hour level. The data collected on Aug $4^{th}$, 2022 (Thursday) is selected to show the city-wide link delay in one day. Fig.~\ref{fig_tram_hour_distribution} and Fig.\ref{fig_bus_hour_distribution} present link delay distributions at different times for tram and bus, respectively. To simplify the description, three hours in one day are shown, which include 7 AM to 8 AM (morning rush hour), 3 PM to 4 PM (normal afternoon), and 6 PM to 7 PM (evening rush hour). In both figures, the first row is the city-wide ground truth link delay, and the second is for prediction by \textit{ArrivalNet-3}. The average link delay below each map is the mean of all public transport links in Dresden. The comparison between different times in one day indicates that compared with non-peak hours, there is a higher link delay during peak hours. This finding applies to both trams and buses. It also shows that it is reasonable to use whether in the rush hour as a contextual feature in the problem formulation. Comparing the average link delay of ground truth and prediction, \textit{ArrivalNet} can predict the trend of delay variation in one day. From the perspective of application, \textit{ArrivalNet} can be used as a city-wide public transport delay monitor.

\section{Conclusions}\label{conclusion}
In this paper, we propose a two-dimensional temporal variation-based bus/tram ATP model, which is termed as \textit{ArrivalNet}. With the FFT transformation, it decomposes one-dimensional time series into two-dimensional temporal variation, that represents both intra-period and inter-period changes. The useful temporal feature of the two-dimensional tensor can be effectively extracted by vision backbones. Moreover, drawing on the concept of ResNet~\cite{he2016deep}, this two-dimensional temporal variation module can be defined as a basic module, allowing for flexible use and adjustment. Validated by a city-wide dataset from the public transport system in Dresden, \textit{ArrivalNet} demonstrates superiority in multi-step bus/tram ATP compared to baseline methods. It can be used for traveler information systems and public transport management systems.

This study focuses on uncovering periodic regularities in temporal information but does not consider the inherent instability of the time series data itself. Future work will focus on mitigating the impact of non-stationary samples on sequential prediction capabilities.

\section*{Acknowledgments}
The authors would like to thank Dresdner Verkehrsbetriebe AG for providing data for this study through the project - Verbesserung der Verkehrsqualität auf dem Stadtring und der Ost-West-Verbindung in Dresden and Henning Jeske for his help in data extraction.

\small
\bibliographystyle{IEEEtran}
\bibliography{references.bib}

\vspace{-10 mm}
\begin{IEEEbiography}[{\includegraphics[width=1in,height=1.25in,clip,keepaspectratio]{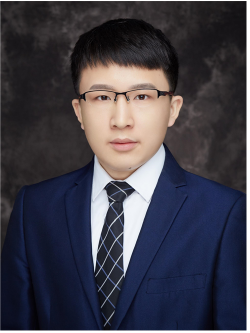}}]{Zirui Li}
received the B.S. degree from the Beijing Institute of Technology (BIT), Beijing, China, in 2019, where he is currently pursuing the Ph.D. degree in mechanical engineering. From June, 2021 to July, 2022, he was a visiting researcher in Delft University of Technology (TU Delft). From Aug, 2022. He was the visiting researcher in the Chair of Traffic Process Automation at the Faculty of Transportation and Traffic Sciences “Friedrich List” of the TU Dresden. His research focuses on interactive behavior modeling, risk assessment and motion planning of automated vehicles.
\end{IEEEbiography}

\begin{IEEEbiography}[{\includegraphics[width=1in,height=1.25in,clip,keepaspectratio]{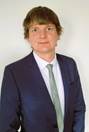}}]{Patrick Wolf}
received the diploma from the Dresden University of Technology (TU Dresden). Dresden, Germany, in 2018, where he is currently pursuing a Ph.D. degree in traffic engineering. From January 2019 to March 2024, he was a researcher in the Chair of Traffic Process Automation at the Faculty of Transportation and Traffic Sciences “Friedrich List” of the Dresden University of Technology (TU Dresden). From April 2024, he is with the traffic management department of the local transportation company (Dresdner Verkehrsbetriebe AG). His research focuses on operational management in public transport, traffic light control and motion forecast of public transport vehicles.
\end{IEEEbiography}

\begin{IEEEbiography}[{\includegraphics[width=1in,height=1.25in,clip,keepaspectratio]{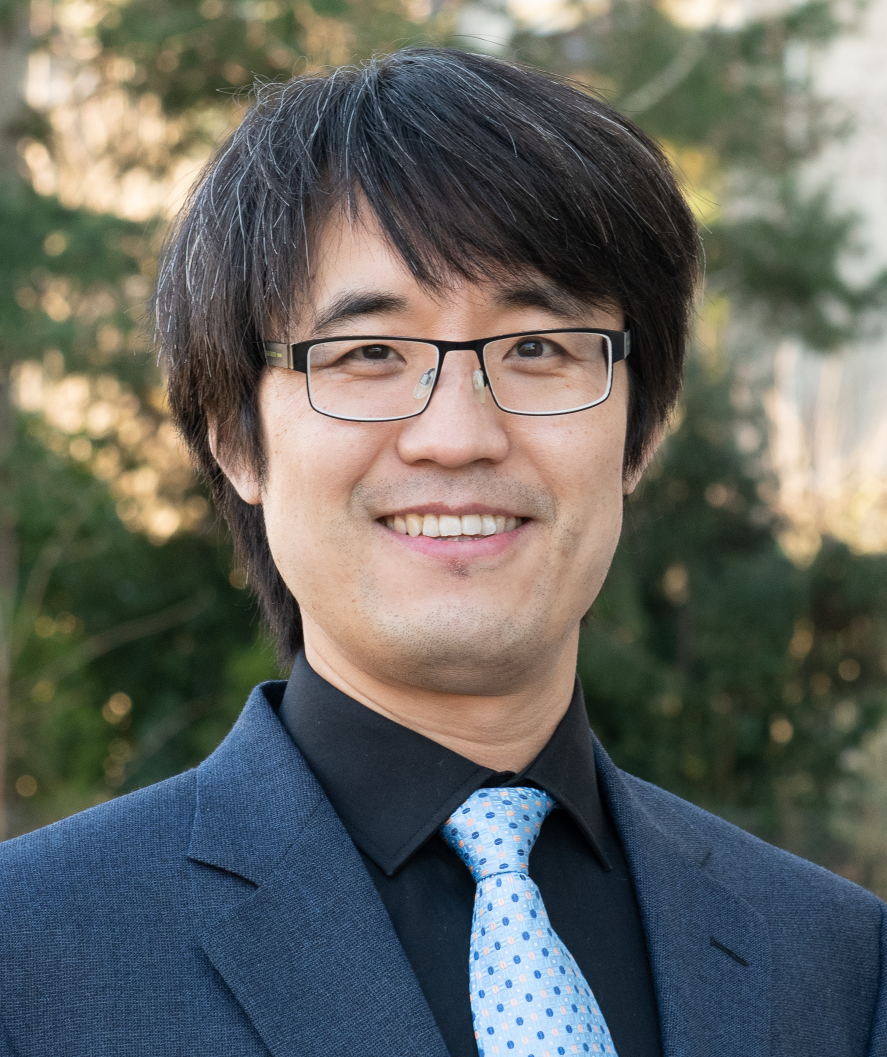}}]{Meng Wang}
received the PhD degree from TU Delft in 2014. He worked as a PostDoc Researcher (2014-2015) at the Faculty of Mechanical Engineering, TU Delft, and as an Assistant Professor (2015-2021) at the Department of Transport and Planning, TU Delft. Since 2021, he has been a Full Professor and Head of the Chair of Traffic Process Automation, “Friedrich List” Faculty of Transport and Traffic Sciences, TU Dresden. His main research interests are control design and impact assessment of Cooperative Intelligent Transportation Systems. He is an Associate Editor of IEEE Transactions on Intelligent Transportation Systems and Transportmetrica B.
\end{IEEEbiography}

\end{document}